\documentclass{article} %
\usepackage[preprint]{colm2026_conference}

\usepackage[table,xcdraw]{xcolor}
\usepackage{xspace}
\usepackage{array}    %
\usepackage{subfig}
\usepackage{caption}

\renewcommand{\arraystretch}{1.05}

\newlength\savewidth
\usepackage{multirow}

\newcommand\tablestyle[2]{%
  \setlength{\tabcolsep}{#1}%
  \renewcommand{\arraystretch}{#2}%
}

\newcommand{\cmark}{\textcolor{green!60!black}{\checkmark}}
\newcommand{\xmark}{\textcolor{red!70!black}{\ding{55}}}
\newcolumntype{x}[1]{>{\centering\arraybackslash}p{#1pt}}
\newcolumntype{y}[1]{>{\raggedleft\arraybackslash}p{#1pt}}
\newcolumntype{z}[1]{>{\raggedright\arraybackslash}p{#1pt}}

\makeatletter
\renewcommand\paragraph{\@startsection{paragraph}{4}{\z@}{0.2ex}{-1em}{\normalfont\normalsize\bfseries}}
\makeatother

\usepackage{minitoc}

\definecolor{baselinecolor}{gray}{.9}

\usepackage{wrapfig}
\usepackage{placeins}

\usepackage{fvextra}
\usepackage{textcomp}

\DefineVerbatimEnvironment{PromptBlock}{Verbatim}{
  breaklines=true,
  breakanywhere=true,
  fontsize=\footnotesize,
  numbers=left,
  numbersep=6pt,
  frame=single,
  framesep=3mm,
  rulecolor=\color{black!30},
  baselinestretch=0.95
}

\usepackage[most]{tcolorbox}
\usepackage{fvextra}

\usepackage{microtype}
\usepackage{hyperref}
\usepackage{url}
\usepackage[normalem]{ulem}
\usepackage{booktabs}
\usepackage{graphicx}
\usepackage{amssymb}
\usepackage{amsmath}
\usepackage{cleveref}
\usepackage{siunitx} %
\usepackage{svg}

\newcommand{\benchmark}{MissionBench}
\newcommand{\numModels}{22}

\usepackage{booktabs, adjustbox, pifont, colortbl, makecell, multirow}
\newcommand{\humancontrolsr}{84.4}
\newcommand{\humancontrolmp}{94.2}
\newcommand{\humancontrolosr}{51.1}

\usepackage{lineno}

\definecolor{darkblue}{rgb}{0, 0, 0.5}

\hypersetup{
colorlinks=true, citecolor=darkblue, linkcolor=darkblue, urlcolor=darkblue,
  pdftitle={
    Zero-Shot Mission-Level Evaluation for Aerial MLLM Agents
  },
  pdfauthor={
    Suman Navaratnarajah, Taehyoung Kim, 
    Jona Ruthardt, Ishaan Bhimwal,
    Ryousuke Yamada, Yannik Blei, Wolfram Burgard,
    Yuki M. Asano
  },
  pdfsubject={
    Zero-shot closed-loop evaluation of aerial MLLM agents
  },
  pdfkeywords={
    multimodal large language models, embodied AI, UAV,
    aerial agents, benchmark, closed-loop control
  },
  pdfdisplaydoctitle=true,
}

\title{
Zero-Shot Mission-Level Evaluation for Aerial MLLM Agents 
}

\author{\noindent
    \textbf{Suman Navaratnarajah$^{1,2,}$\thanks{Equal contribution.}} \quad
    \textbf{Taehyoung Kim$^{2,}$\footnotemark[1]} \quad
    \textbf{Jona Ruthardt$^{1}$\footnotemark[1]} \quad
    \textbf{Ishaan Bhimwal$^{2,3}$} \vspace{0.1em} \\
    \textbf{Ryousuke Yamada$^{1,4}$} \quad  
    \textbf{Yannik Blei$^{1}$} \quad
    \textbf{Wolfram Burgard$^{1}$} \quad
    \textbf{Yuki M. Asano$^{1}$}
    \vspace{0.4em} \\
    \textnormal{
    $^{1}$University of Technology Nuremberg \quad
    $^{2}$Fraunhofer IVI \quad
    $^{3}$THWS \quad
    $^{4}$AIST
    }
}

\makeatletter
\renewenvironment{abstract}
  {\vskip.075in\begin{quote}}
  {\par\end{quote}\vskip1ex}
\makeatother

\begin{document}
\doparttoc
\faketableofcontents

\newcommand{\midsepremove}{\aboverulesep = 0.2mm \belowrulesep = 0.1mm}
\newcommand{\midsepdefault}{\aboverulesep = 0.605mm \belowrulesep = 0.984mm}
\midsepremove

\ifcolmsubmission
\linenumbers
\fi

\vspace{-50pt}

\maketitle

\vspace{-1em}

\begin{abstract}
Multimodal Large Language Models (MLLMs) are emerging as core reasoning modules for embodied agents, yet it remains unclear how well general-purpose models can solve long-horizon embodied tasks from a single high-level instruction. We introduce \textbf{MissionBench}, a benchmark for \textit{mission-level} evaluation of MLLMs in aerial 3D environments. It comprises 120 missions across five simulated 3D environments and four task families. Agents must autonomously plan, navigate, and report outcomes using only egocentric observations and their action history, without aerial-specific fine-tuning.
Across \numModels{} open- and closed-source MLLMs, the strongest model succeeds on fewer than 35\% of missions compared to \humancontrolsr\% human performance, highlighting the difficulty of multi-step embodied tasks. 
Despite large variations between model families, we observe gains from scaling, indicating that larger general-purpose models possess stronger zero-shot embodied capabilities.
Our analysis shows that mission-level competence requires coordinating multiple capabilities beyond spatial perception, including multi-step planning and adaptive reasoning. This motivates closed-loop evaluation and highlights both the promise and risk of scaling-driven improvements for embodied AI.

\vspace{0.5em}
\noindent\textbf{Project website:} \url{https://gomtae.github.io/publications/missionbench}

\end{abstract}

\vspace{-4pt}
\label{sec:intro}
\begin{figure}[!ht]
    \centering
    \includegraphics[width=1.0\textwidth]{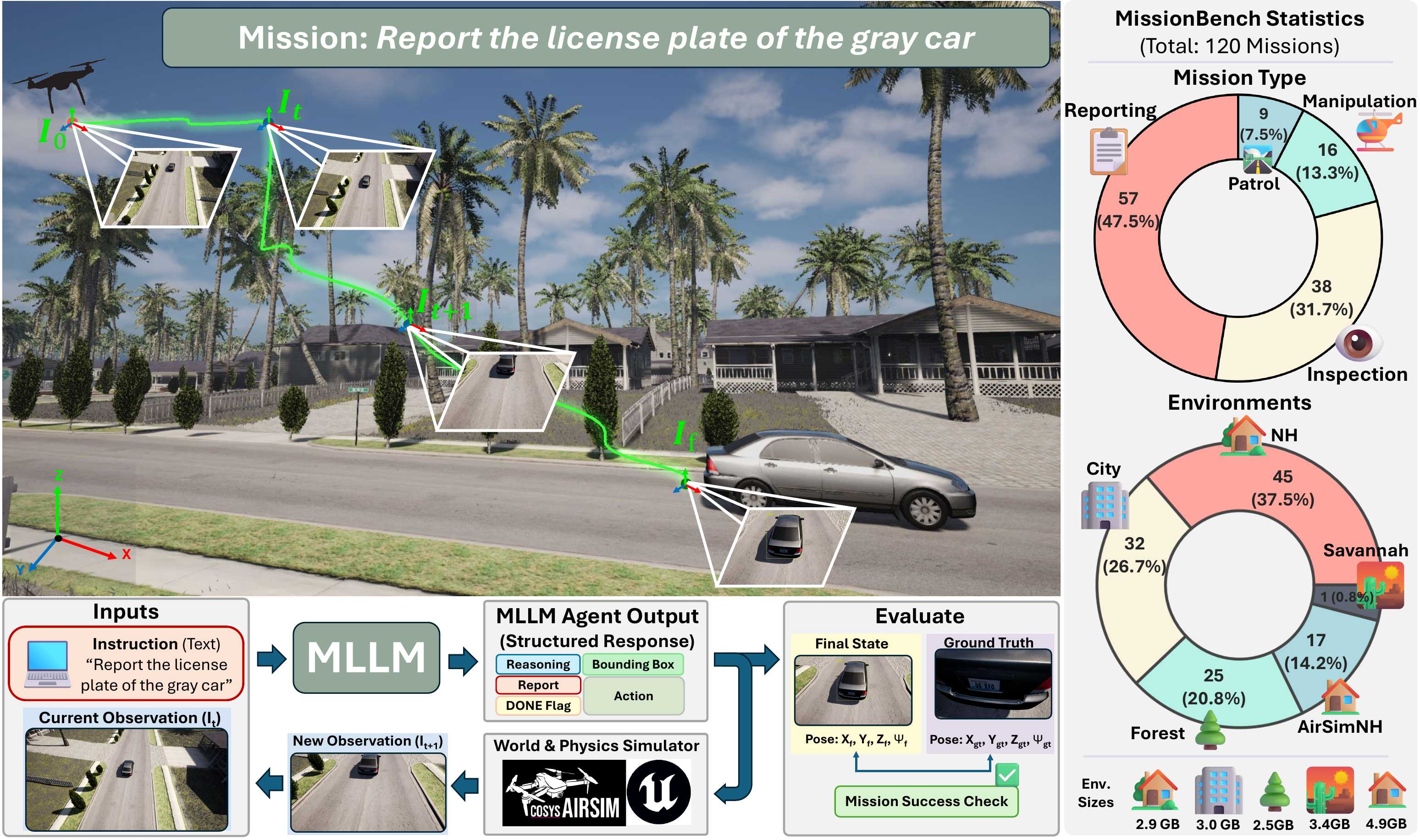}
    \vspace{-16pt}
    \caption{\textbf{\benchmark{}} evaluates MLLMs on mission-level aerial reasoning across four task families: \textbf{Reporting}, \textbf{Inspection}, \textbf{Manipulation}, and \textbf{Patrol}, and five types of environments. Each episode begins with a single natural-language directive. The agent must then iteratively decide where to go, how to position itself, and what results to report to complete the mission.%
    }
    \label{fig:airbench-overview}
\end{figure}

\section{Introduction}

Multimodal Large Language Models (MLLMs) are increasingly used as general-purpose reasoning modules for embodied agents, mapping natural-language instructions and egocentric visual observations to actions~\citep{tian2025uavsmeetllmsoverviews}. This raises a central question: \textit{To what extent can frozen, general-purpose MLLMs execute multi-step embodied missions without task-specific adaptation?} Existing benchmarks provide only partial evidence. Vision-Language Navigation (VLN) typically evaluates whether an agent can follow path-level instructions to reach a destination~\citep{anderson2018vision}, while object navigation benchmarks focus on locating and approaching a target object~\citep{batra2020objectnav}. Task-driven benchmarks such as ALFRED~\citep{shridhar2020alfred} extend beyond navigation, but still rely on step-level subgoals. 
These settings therefore do not directly assess whether frozen MLLMs can coordinate perception, planning, control, and reporting, a critical insight for understanding how progress in general multimodal pretraining translates into embodied mission capability.

Aerial environments provide a particularly demanding testbed for this question because UAV missions are viewpoint-sensitive and open-ended. Hence, success often requires reaching an informative pose, maintaining visual contact, adapting actions over time, and reporting mission-specific outcomes. At the same time, recent work suggests that MLLMs struggle with spatial reasoning, object localization, depth estimation, and relative positioning from aerial viewpoints~\citep{zhang2025your}. Existing aerial benchmarks primarily focus on route following or object search~\citep{gao2025openfly,lee2025citynav,liu2023aerialvln}. \cite{yao2024aeroverse} broaden the evaluation space to include scene reasoning and plan generation, but test these capabilities in isolation. More recent benchmarks consider long-horizon search-and-rescue (SAR) missions or decomposed UAV tasks~\citep{zhang2026esarbench,guo2026bedi}. However, to our knowledge, no prior benchmark jointly evaluates frozen, off-the-shelf MLLMs across multiple mission families with continuous-magnitude pose control, task-specific success criteria, and explicit reporting in a single closed-loop episode.

To address this gap, we introduce \emph{MissionBench}, a benchmark for evaluating frozen, general-purpose MLLMs on mission-level aerial tasks. 
As shown in Fig.~\ref{fig:airbench-overview}, it contains 120 missions distributed across five high-fidelity simulation environments and four task families grounded in real-world Unmanned Aerial Vehicle (UAV) operations: \emph{Reporting} (e.g., ``Read the car's license plate'') \emph{Inspection} (e.g., ``Take a closer look at the burning ship.''), \emph{Manipulation} (e.g., ``Drop a package next to the tent.''), and \emph{Patrol} (e.g., ``Follow the main road.''). Unlike prior benchmarks, each mission is defined by a single natural-language instruction from which the agent must infer \emph{where} to go, \emph{how} to position itself, and \emph{what} to report.

Because aerial tasks are highly sensitive to the viewpoint, \benchmark{} does not restrict movement to fixed-distance primitives (e.g., 3m or 9m steps). Instead, the agent predicts both the direction and magnitude of each action, giving it much finer control over its viewpoint selection. We also introduce a closed-loop evaluation framework that assesses not only how the agent moves, but also whether its actions accomplish the mission.

We evaluate a range of state-of-the-art off-the-shelf MLLMs on \benchmark{} and find that zero-shot drone mission control remains highly challenging. At the same time, performance improves consistently with model scale, suggesting that general-purpose pretraining alone can meaningfully enhance zero-shot embodied capabilities.
These trends point to both an opportunity and a risk: as scaling continues to improve performance, general-purpose MLLMs may become increasingly deployable in real-world systems without task-specific training, raising concerns about reliability, safety, and uncontrolled capability emergence.

Our contributions are:

\begin{enumerate} 

\item \textbf{\benchmark{}}, %
a benchmark of 120 mission-level aerial tasks across five high-fidelity environments and four task families, each defined by a single high-level instruction requiring integrated navigation, viewpoint selection, and reporting.

\item \textbf{A closed-loop evaluation framework} 
that disentangles mission understanding from spatial execution, enabling detailed diagnosis of embodied reasoning failures.

\item \textbf{Empirical analysis of \numModels{} MLLMs}, %
showing a large human-model performance gap and that successful mission completion requires multifaceted capabilities. %

\end{enumerate} 

\section{Related Work}
\label{sec:related}

\begin{table*}[b!]
\centering
\begin{adjustbox}{max width=\textwidth}
\begin{tabular}{@{}l c c c c c c c c c c c c c@{}}
\toprule
\textbf{Benchmark}
  & \textbf{Paradigm}
  & \textbf{RGB-only}
  & \textbf{Action}
  & \textbf{DOF}
  & \textbf{Pose}
  & \textbf{Success}
  & \textbf{Rep}
  & \textbf{Insp}
  & \textbf{Manip}
  & \textbf{Patrol}
  & \textbf{Simulator} \\
\midrule
AerialVLN %
  & Route & \xmark & Disc. & 4 & \xmark & Dist.
  & \xmark & \xmark & \xmark & \xmark
  & AirSim + UE4 \\

AVDN %
  & Route & \cmark & Wpt. & 3 & \xmark & Dist.
  & \xmark & \xmark & \xmark & \xmark
  & xView \\

OpenUAV %
  & Route & \xmark & Wpt. & 6 & \xmark & Dist.
  & \xmark & \xmark & \xmark & \xmark
  & AirSim + UE4 \\

OpenFly %
  & Route & \cmark & Disc. & 4 & \xmark & Dist.
  & \xmark & \xmark & \xmark & \xmark
  & \makecell[c]{AirSim + UE, GTAV,\\[-1pt]GEarth, 3DGS} \\

CityNav %
  & Goal & \xmark & Disc. & 4 & \xmark & Dist.
  & \xmark & \xmark & \xmark & \xmark
  & SensatUrban + Potree \\

\midrule
FlySearch%
  & ObjNav & \cmark & Cont. & 3 & \xmark & Dist.
  & \xmark & \xmark & \xmark & \xmark
  & UnrealCV + UE5 \\

UAV-ON%
  & ObjNav & \xmark & Cont. & 4 & \xmark & Dist.
  & \xmark & \xmark & \xmark & \xmark
  & AirSim + UE \\

\midrule
ESARBench%
  & \makecell[c]{{Mission}} & \xmark & Disc./Wpt. & 4 & \xmark & Mission
  & \xmark & \cmark & \xmark & \xmark
  & AirSim + UE5 \\ %
BEDI%
  & \makecell[c]{{Mission}} & \cmark & Disc./Wpt. & 4 & \cmark & Hierarch.
  & \xmark & \cmark & \cmark & \xmark %
  & AirSim + UE4 \\ %
\rowcolor{blue!6}
\makecell[l]{\textbf{\benchmark{}}\\[-1pt]\textbf{(Ours)}}
  & \textbf{Mission} & \cmark & \textbf{Cont.} & \textbf{4}
  & \cmark & \textbf{Mission}
  & \cmark & \cmark & \cmark & \cmark
  & \textbf{UE5 + CosysAirSim} \\

\bottomrule
\end{tabular}%
\end{adjustbox}
\caption{
\textbf{Comparison of aerial navigation and mission-execution benchmarks.}
\textit{Paradigm:} Route (step-by-step instructions), Goal (destination only), ObjNav (find object/SAR), Mission (task-level objective).
\textit{RGB-only:} Agent relies only on RGB input (no depth or additional sensors).
\textit{Action:} Disc. (discrete primitives), Wpt. (waypoints), Cont. (motion actions with continuously variable magnitude).
\textit{DOF:} Degrees of freedom in control.
\textit{Pose:} Agent must select a viewpoint to complete the task.
\textit{Success:} Dist. (agent/target distance to goal), Hierarch. (hierarchical step-, loop-, and task-level evaluation), Mission (task completion).
\textit{Tasks:} Rep (reporting), Insp (inspection), Manip (manipulation), Patrol.
}
\label{tab:benchmark_comparison}
\end{table*}

\paragraph{Frozen MLLMs as Embodied Reasoning Agents.}
\label{ssec:MLLMasembodied}

MLLMs and LLMs are increasingly used as reasoning backbones for embodied agents. In ground-level robotics, systems such as SayCan \citep{ahn2022can} and PaLM-E \citep{driess2023palm} demonstrate how language models can act as high-level planners grounded in physical affordances. More recent work extends this paradigm to embodied navigation, using MLLMs for instruction-grounded action generation and navigation reasoning \citep{zhang2024navid, zhou2024navgpt, zheng2024towards, cheng2024navila}.
In aerial settings, early approaches use LLMs as task translators with separate vision modules. For example, TypeFly \citep{chen2023typefly} converts natural-language tasks into executable drone programs, and NEUSIS \citep{cai2025neusis} combines MLLM-based perception with neuro-symbolic planning for UAV search. More recent work integrates MLLMs more directly into control. See-Point-Fly \citep{hu2025see} predicts waypoints from images without training, while SkyAgentX \citep{yao2024aeroverse}, OpenFly-Agent \citep{gao2025openfly}, and OpenUAV \citep{wang2024towards} learn aerial decision-making from data. Despite these advances, prior work primarily focuses on navigation. In contrast, \benchmark{} evaluates frozen MLLMs zero-shot on full missions that require viewpoint selection, observation, and reporting.

\paragraph{Aerial Vision-Language Benchmarks.}
\label{ssec:AVLNbenchmarks}

Aerial benchmarks have largely followed the instruction-following template established by indoor benchmarks such as R2R \citep{anderson2018vision} and ALFRED \citep{shridhar2020alfred}.
Most aerial benchmarks focus on path-level navigation and only provide long, step-wise route descriptions. Early work such as AerialVLN \citep{liu2023aerialvln} and AVDN \citep{fan2023aerial} introduced large-scale trajectory datasets, while OpenUAV \citep{wang2024towards} extended this to 6-DOF waypoint prediction. Subsequent efforts, including CityNav \citep{lee2025citynav} and OpenFly \citep{gao2025openfly}, improved realism, scale, and environmental diversity. Despite advances, these works only execute path-level instructions via discrete actions and measure success by reaching a target distance threshold.
A smaller set of benchmarks moves beyond point-to-point navigation. FlySearch~\citep{pardyl2025flysearch} and UAV-ON ~\citep{xiao2025uav} evaluate object-level search and navigation, where the agent must locate a target entity in an aerial scene. ESARBench~\citep{zhang2026esarbench} extends object search to long-horizon (SAR) missions involving clue discovery and victim localization, while BEDI~\citep{guo2026bedi} evaluates heterogeneous UAV tasks through decomposed perception-decision-action loops and hierarchical subtask evaluation.

\paragraph{Beyond aerial navigation.}  Several recent benchmarks evaluate aerial embodied intelligence through isolated sub-skills such as spatial reasoning or perception \citep{yao2024aeroverse, zhang2025your} and show that MLLMs struggle much more in aerial settings than in ground-level ones. This is consistent with broader evidence that MLLMs trained on web-crawled imagery remain weak at spatial reasoning, even in canonical viewpoints \citep{cheng2024spatialrgpt, tong2024cambrian}. However, as they test each capability in isolation, these benchmarks cannot indicate whether model failures compound when an agent must chain perception, planning, and reporting within a single mission. 

As \Cref{tab:benchmark_comparison} summarizes, \benchmark{} uniquely combines continuous 4-DoF control, mission-level success criteria, and explicit reporting in a single closed-loop framework.

\section{\benchmark{}}

\subsection{Task Formulation}
\label{ssec:task}

\begin{figure}[t]
    \centering
    \includegraphics[width=1.0\textwidth]{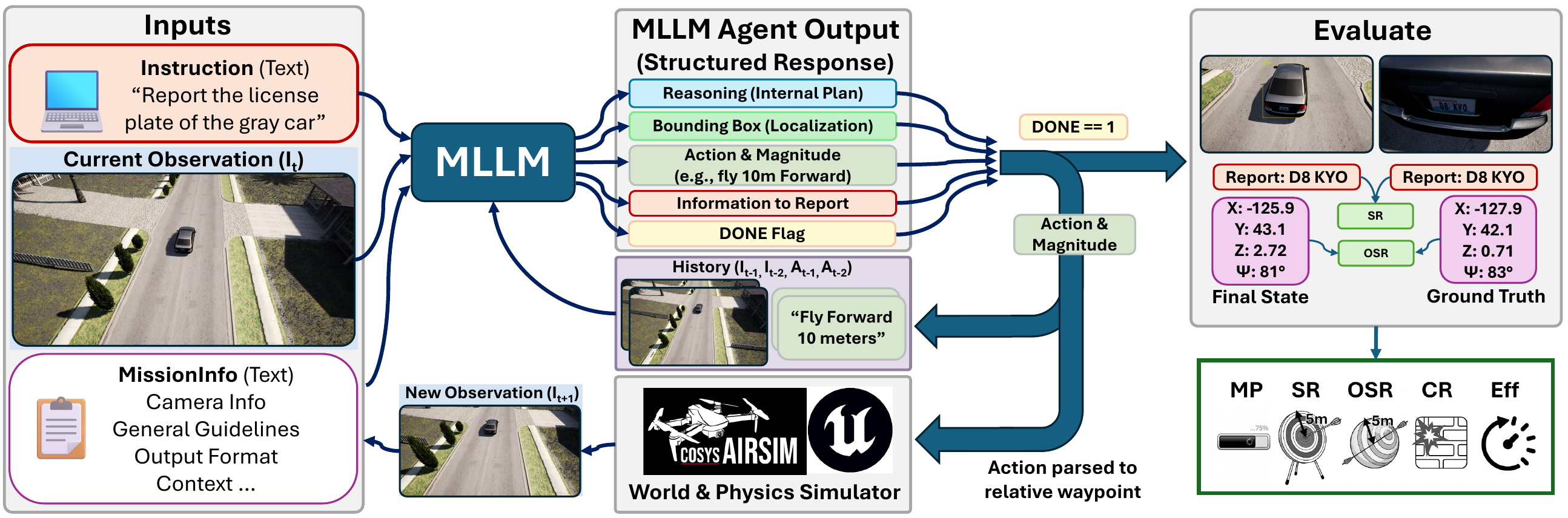}
    \caption{\textbf{The \benchmark{} closed-loop evaluation pipeline}. The MLLM receives a mission instruction and egocentric images and outputs a structured response; actions are parsed into executable waypoints until the agent signals completion. The evaluation metrics Mission Progress (MP), Success Rate (SR), Oracle Success Rate (OSR), Collision Rate (CR) and Step Efficiency (Eff) are described in \ref{ssec:metrics}}
    \label{fig:pipeline}
\end{figure}

We formulate \benchmark{} as an embodied instruction-following problem in which an MLLM (the \textit{agent}) controls a UAV to execute a mission specified via natural language. The overall closed-loop pipeline is visualized in \Cref{fig:pipeline}. At the beginning of each mission, the agent receives a mission instruction $\mathcal{M}$ (e.g., ``Report the license plate number of the red car''), general task information, and an initial egocentric RGB image $I_0$ at pose $\mathbf{p}_0 = (x_0, y_0, z_0, \psi_0)$, where $(x, y, z)$ denotes 3D position in the world frame and $\psi$ the yaw.

The agent then produces a structured response that includes a \textbf{bounding box coordinate} for the relevant region, its \textbf{reasoning} motivating the next step, an \textbf{action primitive and its magnitude}, any mission-relevant \textbf{information to report} (e.g., a license plate number), and a \textbf{binary flag} (\texttt{DONE}) indicating mission completion. While bounding boxes and reasoning traces are ignored in the final evaluation, they provide insights into the agent's intentions and guide the planning akin to chain-of-thought reasoning \citep{cot_paper}. The agent can select from eight directional primitives (forward, backward, strafe left/right, ascend, descend, turn left/right) and move in each with a permissible magnitude up to $d_{\max}$\, meters or $3 \times d_{\max}$\, degrees.
The selected high-level action (e.g., ``fly 20\,m forward'') is parsed into an executable waypoint, and the updated observation is returned to the agent. At each subsequent step $t$, the agent receives the current image $I_t$, up to two previous images, and the history of past actions. The loop continues until the agent signals \texttt{DONE}, optionally with a textual report $\mathcal{R}$, or the step budget $T_{\max}$ is exhausted. While this interface is shared across all tasks, mission completion criteria vary by type as detailed next.

\subsection{Mission Types}
\benchmark{} comprises four distinct mission types motivated by real-world UAV applications~\citep{shakhatreh2019unmanned,mohsan2023unmanned}, as shown in \Cref{fig:airbench-overview}:  

\noindent{\textbf{(i) Reporting}} reflects reconnaissance missions where relevant information must be extracted from egocentric observations and reported back (e.g., reading text, counting objects, or identifying scene attributes).

\noindent{\textbf{(ii) Inspection}} covers infrastructure inspection and search and rescue scenarios that require the agent to reach an observation-ready configuration with the object or region of interest clearly visible in the egocentric view but does not require reporting information.

\noindent{\textbf{(iii) Manipulation}} mimics applications involving landing, payload delivery, or sample collection. The agent must identify and approach the required spatial location and initiate a physical action upon reaching it. 

\noindent{\textbf{(iv) Patrol}} draws from perimeter security, traffic monitoring, and environmental surveillance. Here, agents follow routes along environmental structures (e.g., roads, coastlines, or facility layouts) and may be required to report observations during flight.

\subsection{Benchmark Construction}

\paragraph{Simulation Suite.}
\benchmark{} is built on Unreal Engine 5~\citep{unrealengine} and uses Cosys-AirSim~\citep{jansen2023cosys} for physics-based UAV simulation. It includes five high-fidelity environments drawn from real-world settings: \emph{Neighborhood}, \emph{City}, \emph{Forest}, \emph{Savannah}, and \emph{AirSimNH}. Together, these environments cover a broad range of natural and urban scenes, with different object densities, lighting conditions, and terrain complexity. The closed-loop design using real-time simulations allows for unconstrained agent movement within the environment, a critical advantage over open-loop planning.

\paragraph{Mission Definition.}
Each mission is manually designed within one of the predefined environments and specified by a natural-language instruction $\mathcal{M}$. Depending on the task, the agent may be asked to inspect a target (e.g., read a license plate or assess a fire), carry out a manipulation action (e.g., land, deliver a payload, or collect a sample), or patrol and monitor a specified route or area. The objects or regions relevant to the mission are explicitly placed within the simulation environment during construction. The benchmark spans a wide spectrum of operating ranges, featuring ground-truth trajectory lengths from 13 m up to 4,171 m, with a median of 93 m and a mean of 260 m to evaluate both short-range and long-horizon missions.
Missions are initialized from a fixed start pose chosen so that the target is partially visible but still requires deliberate planning and execution for successful completion. For example, an inspection mission targeting a forest fire might only show subtle smoke cues in the first frame, and a reporting mission targeting a street sign might present it from a high-altitude vantage point where the text is entirely illegible. In both cases, the agent cannot rely on the initial single static frame. It must first detect these distant cues, actively navigate closer or lower, and dynamically adjust its viewpoint orientation to capture or report the required information. This focuses evaluation on viewpoint refinement and mission completion rather than spending the limited action budget on an unconstrained target search. All missions are created and reviewed by multiple human experts to ensure realism, appropriate difficulty, and solvability.

\paragraph{Ground-Truth Acquisition.}
For each mission, a human UAV operator is presented with the mission instruction $\mathcal{M}$ and tasked with solving it within the same simulation environment. The resulting observations, reference trajectories, and terminal state serve as the ground truth for evaluation. 
Because the long-horizon, open-ended nature of \benchmark{} allows for multiple valid trajectories rather than a single optimal solution, we evaluate using metrics that account for this flexibility.

\subsection{Evaluation Metrics}

\label{ssec:metrics}

\noindent\textbf{Success Rate (SR).} Because \benchmark{} evaluates mission completion rather than proximity alone, success criteria are intentionally task-dependent to reflect mission-specific operation requirements. However, they are consolidated into a single \textit{Success Rate} (SR) metric. 
\textbf{Inspection} missions require the agent’s final position $\mathbf{p}_i$ to satisfy $\| \mathbf{p}_i - \mathbf{p}^* \| < \tau_d$ and the orientation threshold $|\psi_i - \psi^*| < \tau_\psi$ compared to the ground-truth position $\mathbf{p}^*$ ($\tau_d = 5$\,m and $\tau_\psi = 15^\circ$). \textbf{Reporting} missions assess whether the extracted information is correct (soft matching via LLM-as-a-judge; details in Sec.~\ref{apdx:llm_as_judge_details}). While success is granted regardless of the final position, the mission design ensures that the information is not extractable from the initial UAV location, first requiring the agent to navigate to an adequate pose. 
\textbf{Manipulation} missions require both the spatial thresholds and the successful execution of the task-specific action.
\textbf{Patrol} missions compute IoU between a $20\,m$  buffer around the ground-truth and generated trajectories, with success at IoU $> 0.5$. Details are shown in \Cref{apdx:evaluation_metric}.

\noindent\textbf{Oracle Success Rate (OSR).} OSR records whether the agent passes within $\tau_d$ of the target at any point during the episode, regardless of mission completion. Comparing OSR with SR helps distinguish whether failure stems from navigation (low OSR) or last-mile perception and reporting (high OSR, low SR).

\noindent\textbf{Mission Progress (MP).} While SR provides only a binary signal, MP captures continuous progress toward the objective: $\text{MP} = \max\!\bigl(0,\, \sum_{t=1}^{T}(d_{t-1} - d_t)\,/\,d_0\bigr)$, where $d_t$ is the distance to the target at step $t$. For Patrol missions, MP equals the non-thresholded IoU.

\noindent\textbf{Step Efficiency (Eff).} $\text{Eff} = 1 - T_{\text{used}} / T_{\max}$, where higher values indicate fewer steps used. Together with MP, it helps separate models that terminate prematurely from those that use up their budget through unproductive exploration.

\noindent\textbf{Collision Rate (CR).} CR is the fraction of steps where a collision occurred, reflecting the MLLM's spatial awareness independently of task success.

\section{Experiments}
\label{sec:experiments}

\begin{table}[t]
\centering
\raggedright
{\scriptsize N=Neighborhood, F=Forest, C=City, S=Savannah, NH=AirSimNH. R=Reporting, I=Inspection, M=Manipulation, P=Patrol.}
\resizebox{\columnwidth}{!}{%
\begin{tabular}{lccll}
\toprule
\textbf{Split} & \textbf{Missions} & \textbf{Environments} & \textbf{Composition} & \textbf{Purpose} \\
\midrule
Proxy                      & 10 & N        & Mixed types             & Spatial perception (\S\ref{ssec:proxy}) \\
Ablation & 5  & N       & Easiest $\subset$ Proxy & Framework ablation (\S\ref{ssec:ablation}) \\
Test                       & 30 & N, F, C & 17\,R / 7\,I / 3\,M / 3\,P & Main evaluation (\S\ref{ssec:full_eval}) \\
Held-out                   & 80 & N, F, C, S, NH              & Remaining               & Future fine-tuning \\
\midrule
\textbf{Total} & \textbf{120} & \textbf{N, F, C, S, NH} & \textbf{57\,R / 38\,I / 16\,M / 9\,P} & \textbf{Overall dataset} \\
\bottomrule
\end{tabular}%
}
\caption{\textbf{Dataset splits.} All splits are disjoint except Ablation, which is a subset of Proxy.}
\label{tab:dataset_split}
\end{table}

\subsection{Experimental Setup}
\label{ssec:setup}

\paragraph{Model setup.} We evaluate \numModels{} open- and closed-source MLLMs with no aerial-specific fine-tuning using their respective APIs (see the full model list in \Cref{tab:model_performance}). 
All models receive the same prompt template, slightly adapted per-mission-type. 
The exact prompts are specified in the Appendix (see \Cref{apdx:prompt_details}).
The visual inputs consist of the three most recent egocentric images at $1920\times1080$ resolution with a fixed camera pitch of $-45^{\circ}$. 
The models use a sampling temperature of $0.7$. The API evaluation cost ranges from \$19.2 (Nova Pro) to \$71.9 (Claude Opus 4.6) for the full Test split; open-weight models can be run locally at the cost of compute (see \Cref{apdx:cost_table}). To establish human performance bounds, subjects evaluated missions either under identical MLLM input/output format constraints (i.e., 3-image history, discrete action budget ($N=2$)) or via continuous real-time keyboard flight control ($N=3$).

\paragraph{Benchmark setup.} The maximum action budget is set to $T_{\max} = 50$ for patrol missions and $T_{\max} = 20$ otherwise. The allowable per-step magnitude $d_{\max}$ (i.e., the maximum distance the agent can traverse in any direction) is automatically chosen to be one fourth of the distance from the start pose to the final pose. This ensures that each mission is solvable well within the maximum allowable time horizon while still permitting flexible control that allows the agent to recover from mistakes and dynamically adjust its step size depending on the target proximity.
Each mission is executed three times per model, and the results are averaged. Episodes are terminated when determined by the model itself or once $T_{\max}$ has been reached. Collisions do not automatically end the episode, but are recorded via the collision rate metric. %

\paragraph{Dataset splits.}
As summarized in \Cref{tab:dataset_split}, we divide the 120 \benchmark{} missions into four subsets, each serving a distinct purpose. The \textit{Proxy} split is used for the perception-focused evaluation (cf.~\Cref{ssec:proxy}), while the \textit{Ablation} split further selects easier missions to study benchmark design choices in a controlled setting with sufficient success signal. The \textit{Test} split is used for the main cross-model evaluation. The remaining 80 missions form the \textit{Held-out} split, which we reserve for future fine-tuning.

\subsection{Performance on \benchmark{}} 
\label{ssec:full_eval}

\Cref{tab:model_performance} reports the main results across all \numModels{} models on the 30 mission test split.  

\begin{table}[t!]
\centering
\vspace{2pt}
\footnotesize
\setlength{\tabcolsep}{4pt}
\begin{tabular}{@{}l rr rrr@{}}
\toprule
\textbf{Model}
  & \textbf{SR\,(\%)}$\uparrow$
  & \textbf{MP}$\uparrow$
  & \textbf{OSR\,(\%)}$\uparrow$
  & \textbf{ CR\,(\%)}$\downarrow$
  & \textbf{Eff\,(\%)}$\uparrow$ \\
\midrule
\rowcolor{black!7}
Random baseline
  & $0.0$ & $5.7$
  & $3.3$ & $0.1$ & $64.8$ \\  
\midrule
Gemini 3.1 Pro
   & \boldmath $34.8$ & \boldmath $73.3$
  & 17.2 & $4.3$ & $\underline{53.6}$ \\

Gemini 2.5 Pro
   & $8.9$  & $54.2$ & $14.4$ & $2.9$ & $18.2$ \\

Gemini 3 Flash
   & $7.9$ & $54.0$
  & \boldmath $20.2$ & $4.9$ & $30.8$ \\

Gemini 2.5 Flash
   & $5.6$ & $39.0$
  & $12.4$ & $4.1$  & $24.7$ \\

Gemini 3.1 Flash Lite
  & $6.9$ & $24.0$ 
  & $8.1$ & $2.5$ & $29.9$ \\

Gemini Robotics 1.5
  & $0.0$ & $35.9$ 
  & $6.9$ & $5.8$ & $32.1$\\
Gemini Robotics 1.6
  & $\underline{32.2}$ & $\underline{70.3}$ 
  & $\underline{17.8}$ & $\underline{0.1}$ & $24.2$\\
Gemma-4-31B-IT
  & $26.7$ & $59.3$ 
  & $11.1$ & $1.6$ & $13.6$\\
\midrule
Claude Opus 4.6
   & $3.3$ & $18.9$
  & $10.0$ & $4.7$ & $16.6$ \\
Claude Sonnet 4.6
   & $4.7$ & $31.2$
  & $7.1$ & $2.0$ & $30.2$ \\

\midrule
GPT-5.4
   & $2.0$ & $45.0$
  & $0.0$ & $4.0$ & $17.9$ \\

GPT-5.4 Mini
   & $1.0$ & $20.0$
  & $0.0$ & $5.0$ & $25.7$ \\

\midrule
Nova Premier
   & $1.1$ & $15.0$
  & $5.6$ & $1.7$ & $17.0$ \\

Nova Pro
   & $0.0$ & $15.2$
  & $3.3$ & $2.0$ & \boldmath  $59.0$ \\
\midrule
Qwen 3.5 2B
   & $0.0$ & $18.1$
  & $5.6$ & \boldmath $0.0$ & $48.3$ \\
  
Qwen 3.5 4B
   & $1.1$ & $28.6$
  & $5.6$ & \boldmath $0.0$ & $36.8$ \\
  
Qwen 3.5 9B
   & $3.3$ & $34.8$
  & $7.8$ & $6.2$ & $29.0$ \\
Qwen 3.5 27B
   & $10.0$ & $53.2$
  & $8.9$ & \boldmath $0.0$ & $16.7$ \\

Qwen 3.6 27B
   & $5.6$ & $32.5$
  & $8.9$ & $3.2$ & $10.6$ \\

Qwen 3.5 35B-A3B
   & $2.2$ & $39.1$
  & $7.8$ & $7.5$ & $32.1$ \\
Qwen 3.6 35B-A3B-Q8
   & $7.9$ & $41.5$
  & $7.9$ & $0.2$ & $29.5$ \\
\midrule
InternVL 3.5 14B
  & $1.1$ & $29.2$
  & $6.7$ & $0.2$ & $11.4$ \\

\midrule
Human baseline (w/ VLM interface)
  & $70.0$ & $79.0$
  & $36.7$ & $0.1$ & $36.0$ \\
\rowcolor{black!7}
Human baseline (w/ keyboard control)
  & $\humancontrolsr$ & $\humancontrolmp$
  & $\humancontrolosr$ & $0.0$ & $-$ \\
\midrule
\end{tabular}
\caption{\textbf{Results on \benchmark.} All metrics are averaged across 3 runs on the 30-mission test split. All metrics are averaged across 3 runs. (SR=Success Rate, MP=Mission Progress, OSR=Oracle Success Rate, CR=Collision Rate, Eff=Step Efficiency). For human baselines, participants (1) followed the same input/output format as models ($N=2$) or (2) were allowed to use the keyboard to continuously control the drone ($N=3$). The random baseline samples a random action primitive and magnitude at every step.}
\label{tab:model_performance}
\end{table}

\paragraph{Overall performance.}
Gemini 3.1 Pro achieves the highest success rate (SR, 34.8\%), followed closely by Gemini Robotics 1.6 (32.2\%) and Gemma-4-31B-IT (26.7\%). Most of the remaining models remain below 10\% SR, indicating that mission-level zero-shot aerial control is still highly challenging. Mission Progress (MP) shows a similar pattern: Gemini 3.1 Pro achieves the highest MP (73.3), followed by Gemini Robotics 1.6 (70.3) and Gemma-4-31B-IT (59.3). Even the best model only succeeds on roughly one out of three missions at the strict threshold, confirming that MissionBench remains a substantial challenge for current MLLMs. At the same time, the gap between stronger and weaker models suggests that general-purpose pretraining and model scaling can substantially improve zero-shot embodied performance.
17 of 19 models with non-zero SR use over 65\% of the available mission-dependent action budget on average ($\text{Eff}<35\%$).
While human operators using continuous keyboard control achieve high success rates (\humancontrolsr), forcing human subjects through the exact structured text-and-image API interface used by the MLLMs drops success to $70.0\%$. This shows that while the discrete text-action interface itself reduces performance to some degree, human participants still significantly outperform MLLMs under identical interface conditions ($34.8\%$ vs. $70.0\%$).

\paragraph{Oracle gap.} 

Oracle Success Rate (OSR) records whether the agent ever passes within 5\,m of the target. Because the distance required for task completion varies (e.g., reading a sign may require 8–10\,m, while manipulation demands close proximity), SR can exceed OSR for perception-heavy tasks. Gemini 3.1 Pro (SR = 34.8\%, OSR = 17.2\%, MP = 73.3) is one such case, where high mission progress confirms genuine task understanding rather than a lucky report. The more diagnostic pattern is OSR $>$ SR: the agent reached the target at one point but failed to complete the mission. Gemini 3 Flash (20.2\% \textit{vs.} 7.9\%) and Claude Opus 4.6 (10.0\% \textit{vs.}\ 3.3\%) both illustrate this gap. Their low MP suggests the agent either drifts away after passing the target or terminates prematurely rather than failing at a final perception step.
GPT-5.4 shows a third pattern with moderate MP (45.0) but near-zero SR and OSR, indicating that the agent makes decent progress toward the goal but never fully converges on the target or fails at the final perception step.

\paragraph{Scaling behavior.}
Mission performance generally improves with scale within model families. For example, Gemini~3.1~Pro substantially outperforms Flash~Lite (73.3 \textit{vs.} 24.0 MP), and GPT-5.4 outperforms GPT-5.4 Mini (45.0 \textit{vs.} 20.0 MP), indicating that larger models exhibit stronger zero-shot embodied capabilities. Within the Qwen~3.5 dense family, scaling is also consistent: increasing model size from 2B to 27B steadily improves MP from 18.1 to 53.2 and SR from 0.0\% to 10.0\%. However, this trend is not strictly monotonic: within the Claude family, Sonnet~4.6 outperforms Opus~4.6 on both MP and SR, and Qwen~3.6 variants do not uniformly improve over their Qwen~3.5 counterparts. More recent model versions, such as Gemini Robotics~1.6 relative to 1.5, also generally exhibit stronger overall performance. This suggests that while scaling is a strong driver of performance, architectural and training differences also play a role.

\paragraph{Spatial awareness.}
The Collision Rate (CR) measures how well each model can infer scene geometry from egocentric visual input when choosing its next action. Collision rates vary from 0.0\% for Qwen 3.5 27B to 7.5\% for Qwen 3.5-35B-A3B. Notably, a stronger task performance does not necessarily mean fewer collisions: Gemini 3.1 Pro, the best overall performer, still records a CR of 4.3\%, which is above the median.

\subsection{Analysis of Perception Abilities}
\label{ssec:proxy}

We test the hypothesis that the performance on \benchmark{} can be largely explained by a model’s single-frame 3D spatial perception abilities. To this end, we evaluate MLLMs in an isolated perception setting that removes navigation and planning. Each model is presented with the initial image $I_0$ of missions in the Proxy split (\Cref{tab:dataset_split}) and tasked with (i) predicting a bounding box for the target object and (ii) estimating its distance (see \Cref{apdx:prompt_details}). We report the mean Intersection over Union (mIoU) for localization and root mean square error (RMSE, in meters) for distance estimation. Results are shown in \Cref{tab:perception}.

The performance on this perception proxy is only a modest-to-moderate association with end-to-end \benchmark{} performance (Pearson: $r_{\text{mIoU,MP}}$ = 0.463, $r_{\text{RMSE,MP}}$ = -0.315). Gemini Robotics 1.6 partly supports this trend, achieving the highest mIoU and relatively strong \benchmark{} performance. However, this relationship is not consistent across metrics or models: Claude Opus 4.6 achieves the lowest RMSE but underperforms on \benchmark{}, while Gemini 2.5 Pro reaches non-trivial MP and SR despite near-zero mIoU. These results suggest that static perception accuracy alone is not sufficient to explain mission-level success.

These results suggest that \benchmark{} captures more than single-frame spatial perception. Success requires integrating perception with instruction following, spatial reasoning, and sequential decision-making over time. Thus, while perception is necessary, it is not sufficient and is unlikely to be the primary bottleneck for current MLLMs.

\subsection{Qualitative Failure Mode Analysis} 
\label{ssec:failure_mode_analysis}

\begin{figure}[t]
    \centering
    \includegraphics[width=1.0\textwidth]{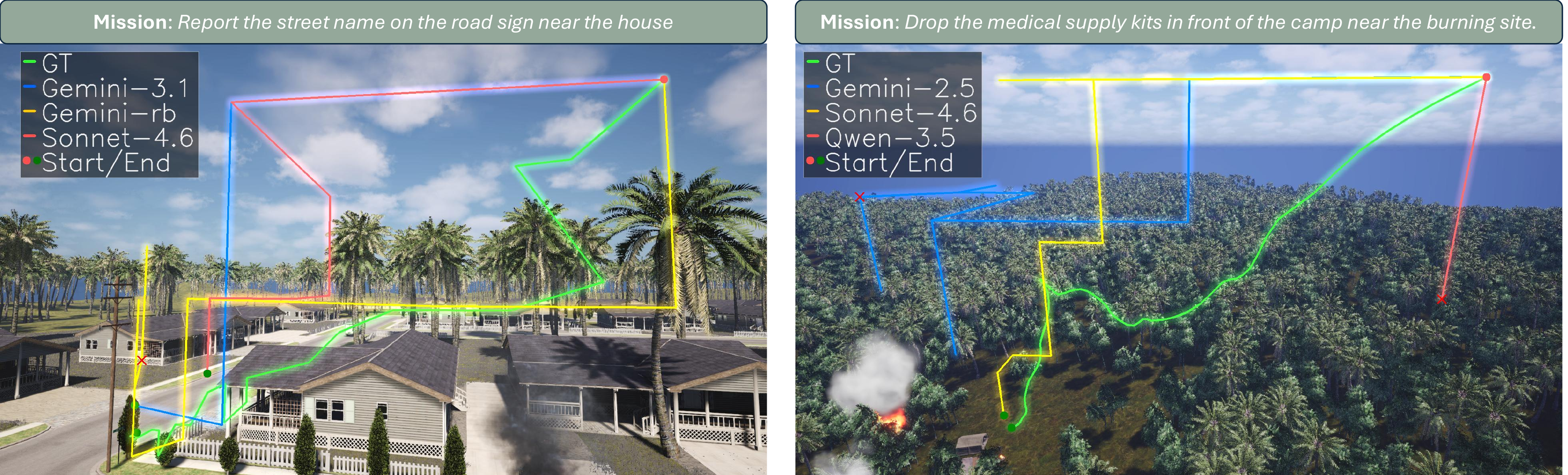}
    \caption{\textbf{Example trajectories.}  The human ground truth (green) is shown together with the trajectories of three MLLMs. The end point of each trajectory is indicated by a green circle (successful mission) or a red cross (failed mission). We examine the causes of these failure cases in Section \ref{apdx:failure_mode_analysis}.
    }
    \label{fig:qualititative_failure_cases}
\end{figure}
The metrics in \Cref{tab:model_performance} quantify \emph{how much} each model achieves. Next, we examine \emph{how} they fail. Step efficiency paired with mission progress provides an initial behavioral grouping. Nova Pro (Eff = 58.96, MP = 15.2) shows high efficiency but low-to-moderate MP, suggesting possible premature termination. Claude Opus~4.6 (Eff = 16.6, MP = 18.9) and Nova Premier (Eff = 17.0, MP = 15.0) consume most of their step budget yet achieve similarly low MP, hinting at aimless exploration or oscillation rather than goal-directed planning. Gemini 3.1 Pro (Eff = 53.6, MP = 73.3) occupies a distinct regime: it uses its budget deliberately and converts that into the highest SR, indicating coherent planning across the episode rather than mere persistence.

We identify three commonly recurring failure patterns across models:
\emph{premature termination}, where the agent declares the mission complete within the first few steps despite being far from the target (Qwen-3.5 in \Cref{fig:qualititative_failure_cases}, right), 
\emph{drift and oscillation}, where the agent exhausts its step budget
circling near landmarks without converging on the goal (Gemini-RB in \Cref{fig:qualititative_failure_cases}, left; Gemini-2.5, right), and \emph{unmet mission objectives}, corresponding to the OSR $>$ SR gap in \Cref{ssec:full_eval}, where the agent reaches the target vicinity but misidentifies the object or its attributes (e.g., rhino as elephant). Further quantitative breakdowns for the failure modes are provided in \Cref{apdx:failure_mode_analysis}.

\subsection{Framework Ablation}
\label{ssec:ablation}

\begin{table*}[t]
\vspace{-.2em}
\centering

\subfloat[
\textbf{Sampling temperature.} %
\label{tab:z-axis_alignment}
]{
\begin{minipage}[t]{0.45\linewidth}\vspace{0pt}
\begin{center}
\tablestyle{4pt}{1.05}
\begin{tabular}{x{90}x{30}x{30}}
Temperature $T$ & SR & MP \\
\midrule
0.0  & 40.0 & 68.1 \\
\rowcolor{gray!25} 0.7  & \textbf{57.1} & \textbf{68.4} \\
1.0  & 46.7 & 59.8 \\
\end{tabular}
\end{center}
\end{minipage}
} \hspace{2em}%
\subfloat[
\textbf{Input resolution.} %
\label{tab:scale_alignment}
]{
\begin{minipage}[t]{0.45\linewidth}\vspace{0pt}
\begin{center}
\tablestyle{4pt}{1.05}
\begin{tabular}{x{90}x{30}x{30}}
Resolution & SR & MP \\
\midrule
$640\times480$ (SD) & 42.9 & \textbf{75.3} \\
$1280\times720$ (HD) & 50.0 & 72.8 \\
\rowcolor{gray!25} $1920\times1080$ (FHD) & \textbf{57.1} & 68.4 \\
\end{tabular}
\end{center}
\end{minipage}
}
\\[-0.2em]
\vspace{.3em}

\subfloat[
\textbf{History length.} %
\label{tab:regularizer_loss}
]{
\begin{minipage}[t]{0.45\linewidth}\vspace{0pt}
\begin{center}
\tablestyle{4pt}{1.05}
\begin{tabular}{x{90}x{30}x{30}}
\# Past Images & SR & MP \\
\midrule
1  & 28.6 & 33.1 \\
\rowcolor{gray!25} 3  & \textbf{57.1} & 68.4 \\
5  & 30.8 & \textbf{69.6} \\
\end{tabular}
\end{center}
\end{minipage}
}\hspace{2em}%
\subfloat[
\textbf{Required structured output fields.} %
\label{tab:data_scaling}
]{
\begin{minipage}[t]{0.45\linewidth}\vspace{0pt}
\begin{center}
\tablestyle{4pt}{1.05}
\begin{tabular}{x{90}x{30}x{30}}
Model output & SR & MP \\
\midrule
\rowcolor{gray!25} Default & \textbf{57.1} & 68.4 \\
w/o BBox & 42.9 & \textbf{70.2} \\
w/o Reasoning & 20.0 & 58.9 \\
\end{tabular}
\end{center}
\end{minipage}
}
\caption{Ablation of key design choices on the 5-mission \textit{Ablation} split using Gemini~3~Flash. One aspect is ablated at a time and default settings are highlighted in \colorbox{baselinecolor}{gray}.}
\label{tab:ablations}
\end{table*}

We ablate several key design choices on the 5-mission \textit{Ablation} split using Gemini~3~Flash, chosen for its representative behavior and moderate inference cost. Starting from the default setup specified in \Cref{ssec:setup}, we vary one component at a time (\Cref{tab:ablations}). Overall, the ablations suggest that mission success is more sensitive to design choices than mission progress as several variants still achieve comparable MP, but fail to convert that progress into successful completion.

\noindent\textbf{Temperature.}
Moderate stochasticity performs best and the default $T{=}0.7$ yields the highest SR (57.1) with essentially unchanged MP relative to deterministic decoding ($68.4$ vs.\ $68.1$), while a higher temperature ($T{=}1.0$) reduces both SR and MP. This suggests that some sampling helps the agent recover from early mistakes, whereas overly random decoding destabilizes long-horizon behavior.

\noindent\textbf{Resolution.}
Higher resolution mainly improves task completion rather than coarse progress. Reducing the input from FHD to SD increases MP (75.3 vs.\ 68.4) but lowers SR substantially (42.9 vs.\ 57.1), indicating that lower-resolution inputs are often sufficient for moving toward the goal, but not for the fine-grained perception and reporting needed for successful mission completion.

\noindent\textbf{History length.}
Access to previous states is important, but more is not always better. Using only one past image sharply reduces both SR and MP, showing that temporal context helps track progress and recover from mistakes. Increasing the history from three to five images preserves MP (69.6 vs.\ 68.4) but lowers SR markedly (30.8 vs.\ 57.1), suggesting that longer context may dilute the most relevant recent information.

\noindent\textbf{Required structured output fields.}
The structured output is particularly important for reliable mission completion. Removing the reasoning field causes the largest drop in both SR and MP, indicating that explicitly eliciting intermediate reasoning helps sustain coherent behavior. Removing the bounding box lowers SR while keeping MP high, suggesting that explicit target localization is less important for navigation than for the final task completion.

\section{Conclusion}
\label{sec:conclusion}

We presented \benchmark{}, a benchmark for evaluating MLLMs at the mission-level of aerial reasoning. Across \numModels{} models, even the strongest completes roughly one third of all missions, and single-frame spatial perception is a poor predictor of mission success. Our metric suite and failure-mode analysis show that models fail in several distinct ways, including premature termination, aimless drift, and unmet mission objectives. None of these patterns is visible from proxy-task evaluation alone.

\section{Acknowledgements}
The authors gratefully acknowledge the HPC resources provided by the Erlangen National High Performance Computing Center (NHR@FAU) of the Friedrich-Alexander Universität Erlangen-Nürnberg (FAU) under the BayernKI project v115be. BayernKI funding is provided by Bavarian state authorities. %

\newpage

\bibliography{colm2026_conference}

\begin{thebibliography}{33}
\providecommand{\natexlab}[1]{#1}
\providecommand{\url}[1]{\texttt{#1}}
\expandafter\ifx\csname urlstyle\endcsname\relax
  \providecommand{\doi}[1]{doi: #1}\else
  \providecommand{\doi}{doi: \begingroup \urlstyle{rm}\Url}\fi

\bibitem[Ahn et~al.(2022)Ahn, Brohan, Brown, Chebotar, Cortes, David, Finn, Fu, Gopalakrishnan, Hausman, et~al.]{ahn2022can}
Michael Ahn, Anthony Brohan, Noah Brown, Yevgen Chebotar, Omar Cortes, Byron David, Chelsea Finn, Chuyuan Fu, Keerthana Gopalakrishnan, Karol Hausman, et~al.
\newblock Do as i can, not as i say: Grounding language in robotic affordances.
\newblock \emph{arXiv preprint arXiv:2204.01691}, 2022.

\bibitem[Anderson et~al.(2018)Anderson, Wu, Teney, Bruce, Johnson, S{\"u}nderhauf, Reid, Gould, and Van Den~Hengel]{anderson2018vision}
Peter Anderson, Qi~Wu, Damien Teney, Jake Bruce, Mark Johnson, Niko S{\"u}nderhauf, Ian Reid, Stephen Gould, and Anton Van Den~Hengel.
\newblock Vision-and-language navigation: Interpreting visually-grounded navigation instructions in real environments.
\newblock In \emph{Proceedings of the IEEE conference on computer vision and pattern recognition}, pp.\  3674--3683, 2018.

\bibitem[Batra et~al.(2020)Batra, Gokaslan, Kembhavi, Maksymets, Mottaghi, Savva, Toshev, and Wijmans]{batra2020objectnav}
Dhruv Batra, Aaron Gokaslan, Aniruddha Kembhavi, Oleksandr Maksymets, Roozbeh Mottaghi, Manolis Savva, Alexander Toshev, and Erik Wijmans.
\newblock Objectnav revisited: On evaluation of embodied agents navigating to objects.
\newblock \emph{arXiv preprint arXiv:2006.13171}, 2020.

\bibitem[Cai et~al.(2025)Cai, Cardenas, Leo, Zhang, Backman, Li, Li, Ghorbanali, Datta, Qu, et~al.]{cai2025neusis}
Zhixi Cai, Cristian~Rojas Cardenas, Kevin Leo, Chenyuan Zhang, Kal Backman, Hanbing Li, Boying Li, Mahsa Ghorbanali, Stavya Datta, Lizhen Qu, et~al.
\newblock Neusis: A compositional neuro-symbolic framework for autonomous perception, reasoning, and planning in complex uav search missions.
\newblock \emph{IEEE Robotics and Automation Letters}, 2025.

\bibitem[Chen et~al.(2023)Chen, Yu, Ling, and Zhong]{chen2023typefly}
Guojun Chen, Xiaojing Yu, Neiwen Ling, and Lin Zhong.
\newblock Typefly: Flying drones with large language model.
\newblock \emph{arXiv preprint arXiv:2312.14950}, 2023.

\bibitem[Cheng et~al.(2024{\natexlab{a}})Cheng, Ji, Yang, Gongye, Zou, Kautz, B{\i}y{\i}k, Yin, Liu, and Wang]{cheng2024navila}
An-Chieh Cheng, Yandong Ji, Zhaojing Yang, Zaitian Gongye, Xueyan Zou, Jan Kautz, Erdem B{\i}y{\i}k, Hongxu Yin, Sifei Liu, and Xiaolong Wang.
\newblock Navila: Legged robot vision-language-action model for navigation.
\newblock \emph{arXiv preprint arXiv:2412.04453}, 2024{\natexlab{a}}.

\bibitem[Cheng et~al.(2024{\natexlab{b}})Cheng, Yin, Fu, Guo, Yang, Kautz, Wang, and Liu]{cheng2024spatialrgpt}
An-Chieh Cheng, Hongxu Yin, Yang Fu, Qiushan Guo, Ruihan Yang, Jan Kautz, Xiaolong Wang, and Sifei Liu.
\newblock Spatialrgpt: Grounded spatial reasoning in vision-language models.
\newblock \emph{Advances in Neural Information Processing Systems}, 37:\penalty0 135062--135093, 2024{\natexlab{b}}.

\bibitem[Driess et~al.(2023)Driess, Xia, Sajjadi, Lynch, Chowdhery, Ichter, Wahid, Tompson, Vuong, Yu, et~al.]{driess2023palm}
Danny Driess, Fei Xia, Mehdi~SM Sajjadi, Corey Lynch, Aakanksha Chowdhery, Brian Ichter, Ayzaan Wahid, Jonathan Tompson, Quan Vuong, Tianhe Yu, et~al.
\newblock Palm-e: An embodied multimodal language model.
\newblock \emph{arXiv preprint arXiv:2303.03378}, 2023.

\bibitem[{Epic Games}(2019)]{unrealengine}
{Epic Games}.
\newblock Unreal engine, 2019.
\newblock URL \url{https://www.unrealengine.com}.

\bibitem[Fan et~al.(2023)Fan, Chen, Jiang, Zhou, Zhang, and Wang]{fan2023aerial}
Yue Fan, Winson Chen, Tongzhou Jiang, Chun Zhou, Yi~Zhang, and Xin Wang.
\newblock Aerial vision-and-dialog navigation.
\newblock In \emph{Findings of the Association for Computational Linguistics: ACL 2023}, pp.\  3043--3061, 2023.

\bibitem[Gao et~al.(2025)Gao, Li, You, Liu, Li, Chen, Chen, Tang, Wang, Yang, et~al.]{gao2025openfly}
Yunpeng Gao, Chenhui Li, Zhongrui You, Junli Liu, Zhen Li, Pengan Chen, Qizhi Chen, Zhonghan Tang, Liansheng Wang, Penghui Yang, et~al.
\newblock Openfly: A comprehensive platform for aerial vision-language navigation.
\newblock \emph{arXiv preprint arXiv:2502.18041}, 2025.

\bibitem[Guo et~al.(2026)Guo, Wu, He, Li, Li, and Tao]{guo2026bedi}
Mingning Guo, Mengwei Wu, Jiarun He, Shaoxian Li, Haifeng Li, and Chao Tao.
\newblock Bedi: A comprehensive benchmark for evaluating embodied agents on uavs.
\newblock \emph{ISPRS Journal of Photogrammetry and Remote Sensing}, 232:\penalty0 910--936, 2026.

\bibitem[Hu et~al.(2025)Hu, Lin, Lee, Su, Lee, Tsai, Lin, Chen, Ke, and Liu]{hu2025see}
Chih~Yao Hu, Yang-Sen Lin, Yuna Lee, Chih-Hai Su, Jie-Ying Lee, Shr-Ruei Tsai, Chin-Yang Lin, Kuan-Wen Chen, Tsung-Wei Ke, and Yu-Lun Liu.
\newblock See, point, fly: A learning-free vlm framework for universal unmanned aerial navigation.
\newblock In \emph{Conference on Robot Learning}, pp.\  4697--4708. PMLR, 2025.

\bibitem[Jansen et~al.(2023)Jansen, Verreycken, Schenck, Blanquart, Verhulst, Huebel, and Steckel]{jansen2023cosys}
Wouter Jansen, Erik Verreycken, Anthony Schenck, Jean-Edouard Blanquart, Connor Verhulst, Nico Huebel, and Jan Steckel.
\newblock Cosys-airsim: a real-time simulation framework expanded for complex industrial applications.
\newblock \emph{Annual Modeling and Simulation Conference (ANNSIM)}, 2023.

\bibitem[Lee et~al.(2025)Lee, Miyanishi, Kurita, Sakamoto, Azuma, Matsuo, and Inoue]{lee2025citynav}
Jungdae Lee, Taiki Miyanishi, Shuhei Kurita, Koya Sakamoto, Daichi Azuma, Yutaka Matsuo, and Nakamasa Inoue.
\newblock Citynav: A large-scale dataset for real-world aerial navigation.
\newblock In \emph{Proceedings of the IEEE/CVF International Conference on Computer Vision}, pp.\  5912--5922, 2025.

\bibitem[Liu et~al.(2023)Liu, Zhang, Qi, Wang, Zhang, and Wu]{liu2023aerialvln}
Shubo Liu, Hongsheng Zhang, Yuankai Qi, Peng Wang, Yanning Zhang, and Qi~Wu.
\newblock Aerialvln: Vision-and-language navigation for uavs.
\newblock In \emph{Proceedings of the IEEE/CVF International Conference on Computer Vision}, pp.\  15384--15394, 2023.

\bibitem[Mohsan et~al.(2023)Mohsan, Othman, Li, Alsharif, and Khan]{mohsan2023unmanned}
Syed Agha~Hassnain Mohsan, Nawaf Qasem~Hamood Othman, Yanlong Li, Mohammed~H Alsharif, and Muhammad~Asghar Khan.
\newblock Unmanned aerial vehicles (uavs): Practical aspects, applications, open challenges, security issues, and future trends.
\newblock \emph{Intelligent service robotics}, 16\penalty0 (1):\penalty0 109--137, 2023.

\bibitem[Pardyl et~al.(2025)Pardyl, Matuszek, Przebieracz, Cygan, et~al.]{pardyl2025flysearch}
Adam Pardyl, Dominik Matuszek, Mateusz Przebieracz, Marek Cygan, et~al.
\newblock Flysearch: Exploring how vision-language models explore.
\newblock \emph{NeurIPS Datasets and Benchmarks}, 2025.

\bibitem[Qiu et~al.(2017)Qiu, Zhong, Zhang, Qiao, Xiao, Kim, and Wang]{qiu2017unrealcv}
Weichao Qiu, Fangwei Zhong, Yi~Zhang, Siyuan Qiao, Zihao Xiao, Tae~Soo Kim, and Yizhou Wang.
\newblock Unrealcv: Virtual worlds for computer vision.
\newblock In \emph{Proceedings of the 25th ACM international conference on Multimedia}, pp.\  1221--1224, 2017.

\bibitem[Shah et~al.(2017)Shah, Dey, Lovett, and Kapoor]{shah2017airsim}
Shital Shah, Debadeepta Dey, Chris Lovett, and Ashish Kapoor.
\newblock Airsim: High-fidelity visual and physical simulation for autonomous vehicles.
\newblock In \emph{Field and service robotics: Results of the 11th international conference}, pp.\  621--635. Springer, 2017.

\bibitem[Shakhatreh et~al.(2019)Shakhatreh, Sawalmeh, Al-Fuqaha, Dou, Almaita, Khalil, Othman, Khreishah, and Guizani]{shakhatreh2019unmanned}
Hazim Shakhatreh, Ahmad~H Sawalmeh, Ala Al-Fuqaha, Zuochao Dou, Eyad Almaita, Issa Khalil, Noor~Shamsiah Othman, Abdallah Khreishah, and Mohsen Guizani.
\newblock Unmanned aerial vehicles (uavs): A survey on civil applications and key research challenges.
\newblock \emph{IEEE access}, 7:\penalty0 48572--48634, 2019.

\bibitem[Shridhar et~al.(2020)Shridhar, Thomason, Gordon, Bisk, Han, Mottaghi, Zettlemoyer, and Fox]{shridhar2020alfred}
Mohit Shridhar, Jesse Thomason, Daniel Gordon, Yonatan Bisk, Winson Han, Roozbeh Mottaghi, Luke Zettlemoyer, and Dieter Fox.
\newblock Alfred: A benchmark for interpreting grounded instructions for everyday tasks.
\newblock In \emph{Proceedings of the IEEE/CVF conference on computer vision and pattern recognition}, pp.\  10740--10749, 2020.

\bibitem[Tian et~al.(2025)Tian, Lin, Li, Zhang, Zhang, Fu, Huang, Dai, Wang, Tian, Li, Lv, Kovács, and Wang]{tian2025uavsmeetllmsoverviews}
Yonglin Tian, Fei Lin, Yiduo Li, Tengchao Zhang, Qiyao Zhang, Xuan Fu, Jun Huang, Xingyuan Dai, Yutong Wang, Chunwei Tian, Bai Li, Yisheng Lv, Levente Kovács, and Fei-Yue Wang.
\newblock {UAVs} meet {LLMs}: Overviews and perspectives toward agentic low-altitude mobility, 2025.

\bibitem[Tong et~al.(2024)Tong, Brown, Wu, Woo, Middepogu, Akula, Yang, Yang, Iyer, Pan, et~al.]{tong2024cambrian}
Shengbang Tong, Ellis Brown, Penghao Wu, Sanghyun Woo, Manoj Middepogu, Sai~C Akula, Jihan Yang, Shusheng Yang, Adithya Iyer, Xichen Pan, et~al.
\newblock Cambrian-1: A fully open, vision-centric exploration of multimodal llms.
\newblock \emph{Advances in Neural Information Processing Systems}, 37:\penalty0 87310--87356, 2024.

\bibitem[Wang et~al.(2025)Wang, Yang, Wang, Kwan, Chen, Wu, Li, Liao, and Liu]{wang2024towards}
Xiangyu Wang, Donglin Yang, Ziqin Wang, Hohin Kwan, Jinyu Chen, Wenjun Wu, Hongsheng Li, Yue Liao, and Si~Liu.
\newblock Towards realistic uav vision-language navigation: Platform, benchmark, and methodology.
\newblock \emph{International Conference on Learning Representations (ICLR)}, 2025.

\bibitem[Wei et~al.(2022)Wei, Wang, Schuurmans, Bosma, ichter, Xia, Chi, Le, and Zhou]{cot_paper}
Jason Wei, Xuezhi Wang, Dale Schuurmans, Maarten Bosma, brian ichter, Fei Xia, Ed~Chi, Quoc~V Le, and Denny Zhou.
\newblock Chain-of-thought prompting elicits reasoning in large language models.
\newblock In \emph{Advances in Neural Information Processing Systems}, volume~35, 2022.

\bibitem[Xiao et~al.(2025)Xiao, Sun, Shao, Gan, Liu, Wu, Guan, and Deng]{xiao2025uav}
Jianqiang Xiao, Yuexuan Sun, Yixin Shao, Boxi Gan, Rongqiang Liu, Yanjin Wu, Weili Guan, and Xiang Deng.
\newblock Uav-on: A benchmark for open-world object goal navigation with aerial agents.
\newblock In \emph{Proceedings of the 33rd ACM International Conference on Multimedia}, pp.\  13023--13029, 2025.

\bibitem[Yao et~al.(2024)Yao, Yue, Liu, Sun, and Fu]{yao2024aeroverse}
Fanglong Yao, Yuanchang Yue, Youzhi Liu, Xian Sun, and Kun Fu.
\newblock Aeroverse: Uav-agent benchmark suite for simulating, pre-training, finetuning, and evaluating aerospace embodied world models.
\newblock \emph{arXiv preprint arXiv:2408.15511}, 2024.

\bibitem[Zhang et~al.(2026)Zhang, Chen, Zhou, and Yang]{zhang2026esarbench}
Daoxuan Zhang, Ping Chen, Jianyi Zhou, and Shuo Yang.
\newblock Esarbench: A benchmark for agentic uav embodied search and rescue.
\newblock \emph{arXiv preprint arXiv:2605.01371}, 2026.

\bibitem[Zhang et~al.(2024)Zhang, Wang, Xu, Zhou, Hong, Fang, Wu, Zhang, and Wang]{zhang2024navid}
Jiazhao Zhang, Kunyu Wang, Rongtao Xu, Gengze Zhou, Yicong Hong, Xiaomeng Fang, Qi~Wu, Zhizheng Zhang, and He~Wang.
\newblock Navid: Video-based vlm plans the next step for vision-and-language navigation.
\newblock \emph{arXiv preprint arXiv:2402.15852}, 2024.

\bibitem[Zhang et~al.(2025)Zhang, Zhang, Li, Fu, Tang, Ye, Chen, Liang, Hao, and Ding]{zhang2025your}
Lingfeng Zhang, Yuchen Zhang, Hongsheng Li, Haoxiang Fu, Yingbo Tang, Hangjun Ye, Long Chen, Xiaojun Liang, Xiaoshuai Hao, and Wenbo Ding.
\newblock Is your vlm sky-ready? a comprehensive spatial intelligence benchmark for uav navigation.
\newblock \emph{arXiv preprint arXiv:2511.13269}, 2025.

\bibitem[Zheng et~al.(2024)Zheng, Huang, Zhao, Zhong, and Wang]{zheng2024towards}
Duo Zheng, Shijia Huang, Lin Zhao, Yiwu Zhong, and Liwei Wang.
\newblock Towards learning a generalist model for embodied navigation.
\newblock In \emph{Proceedings of the IEEE/CVF Conference on Computer Vision and Pattern Recognition}, pp.\  13624--13634, 2024.

\bibitem[Zhou et~al.(2024)Zhou, Hong, Wang, Wang, and Wu]{zhou2024navgpt}
Gengze Zhou, Yicong Hong, Zun Wang, Xin~Eric Wang, and Qi~Wu.
\newblock Navgpt-2: Unleashing navigational reasoning capability for large vision-language models.
\newblock In \emph{European Conference on Computer Vision}, pp.\  260--278. Springer, 2024.

\end{thebibliography}
\bibliographystyle{colm2026_conference}

\clearpage
\appendix
\addcontentsline{toc}{section}{Appendices}
\setcounter{table}{0} \setcounter{figure}{0}
\renewcommand{\thetable}{\Alph{section}\arabic{table}}
\renewcommand{\thefigure}{\Alph{section}\arabic{figure}}
\pagenumbering{arabic}

\part{\LARGE{Supplementary Appendix}}

{
\hypersetup{linkcolor=black}
\parttoc
}
\bigskip

In the following, we provide supplementary discussion, analyses, and implementation details for \benchmark{}. The appendix covers limitations and future work (\Cref{apdx:discussion}), extended results (\Cref{apdx:extended_results}), benchmark design (\Cref{apdx:details_on_benchmark_generation}), and model-level information on prompts and inference costs (\Cref{apdx:model_details}).

\section{Further Discussion}
\label{apdx:discussion}
\subsection{Limitations and Future Work}
\label{apdx:limitations}

\benchmark{} provides a controlled evaluation of zero-shot mission execution in high-fidelity simulation and does not imply real-world flight readiness as a sim-to-real gap may persist. Although the 120-mission set is diverse, it remains limited in scale and cannot capture real-world factors such as sensor noise, dynamic obstacles, or adversarial conditions. Success is defined relative to human reference trajectories and task-specific tolerances, while still allowing multiple valid solutions rather than assuming a single optimal path.

We use standardized prompts across models to ensure fair comparison, but per-model prompt optimization may further improve performance and reduce model-specific failure modes.

\benchmark{}’s 120 missions are slightly skewed toward reporting (N=57) and inspection missions (N=38) with fewer manipulation (N=16) and patrol (N=9) missions, reflecting mission-construction complexity and structure.
Patrol missions require large coherent route layouts and environment-level coverage, while manipulation missions add task-specific actions atop target localization. Reporting and inspection missions are more numerous as they cover diverse operational variants (e.g., reading text, counting/identifying/inspecting objects, and reporting attributes), compensating for the imbalance with higher intra-type variations and maintaining diverse benchmark-level scenarios. We report per-type results in \Cref{tab:task_breakdown}.

We reserve the 80-mission held-out split for future fine-tuning experiments to study whether aerial-specific training can reduce the gap between perception and execution. A key next step is extending \benchmark{} toward real-world evaluation.

\subsection{Ethical Considerations and Broader Impacts}
\label{ssec:ethics}

\paragraph{Dual-use and deployment risks.}
Our results suggest that general-purpose MLLMs can exhibit non-trivial zero-shot embodied capabilities without task-specific training. While this enables rapid prototyping and lowers engineering barriers, it also raises concerns about premature deployment in safety-critical systems such as drones. Models that perform well in simulation may behave unpredictably under distribution shift, partial observability, or long-horizon feedback loops.

\paragraph{Practical deployment constraints.}
Despite these capabilities, current MLLMs are far from directly deployable on real-world aerial platforms. They require large-scale compute and typically rely on cloud-based inference, making on-device execution infeasible and introducing latency, reliability, and connectivity constraints. As such, strong benchmark performance should not be interpreted as readiness for real-world deployment.

\paragraph{Uncontrolled capability emergence.}
The observed gains from scaling suggest that embodied capabilities can emerge from general-purpose pretraining rather than explicit design for control. Such behavior is difficult to anticipate, evaluate, or constrain, raising challenges for validation and certification in real-world systems.

\paragraph{Evaluation limitations and over-reliance on benchmarks.}
Although \benchmark{} introduces a closed-loop evaluation framework, it remains a simulated benchmark. Strong performance may therefore overestimate real-world capability. We advocate complementing benchmark results with real-world testing and system-level safety validation.

\paragraph{Potential benefits.}
At the same time, the ability of general-purpose models to generalize to embodied tasks without fine-tuning could lower the barrier to entry for robotics and UAV applications, enabling more flexible systems and reducing reliance on large task-specific datasets.

\subsection{Disclosure of LLM Use}
\label{apdx:llm_use}

During development, the authors used LLM-based tools as coding assistants and for language editing (e.g., paraphrasing and grammar correction).
While evaluating MLLMs, \benchmark{} itself largely employs deterministic and rule-based evaluation metrics. Only for ensuring that the report $\mathcal{R}$ matches the ground truth observation, a LLM-as-a-judge is employed and was found to be highly reliable with manual spot checks on 150 random samples (cf. \Cref{apdx:llm_as_judge_details}).

\section{Extended Results}
\label{apdx:extended_results}

\subsection{Perception Proxy vs.\ Mission-Level Performance}
\label{apdx:proxyvsmission}

\begin{table}[t]
\centering
\vspace{2pt}
\footnotesize
\setlength{\tabcolsep}{4pt}
\begin{tabular}{
  @{}lrrrr@{}
}
\toprule
& \multicolumn{2}{c}{\textbf{Perception Proxy}}
& \multicolumn{2}{c}{\textbf{\benchmark{}}} \\
\cmidrule(rr){2-3} \cmidrule(rr){4-5}
\textbf{Model}
& {\textbf{mIoU}$\uparrow$}
& {\textbf{RMSE\,(m)}$\downarrow$}
& {\textbf{MP}$\uparrow$}
& {\textbf{SR\,(\%)}$\uparrow$} \\
\midrule
Gemini 3.1 Pro         & 0.51          & 27.7          & \bfseries 73.3          & \bfseries 34.8          \\
Gemini 2.5 Pro         & 0.01          & 53.4          & 54.2          & 8.9           \\
Gemini 3 Flash         & 0.58 & 72.3         & 54.0          & 7.9           \\
Gemini 2.5 Flash       & 0.00          & 69.3          & 39.0 & 5.6 \\
Gemini 3.1 Flash Lite  & 0.45          & 71.4          & 24.0 & 6.9 \\
Gemini Robotics 1.5    & 0.03          & 20.0          & 35.9 & 0.0 \\
Gemma-4-31B-IT    & 0.47          & 12.5          & 59.3 & 26.7 \\
Gemini Robotics 1.6    & \bfseries 0.66          & 61.1          & 70.3 & 32.2 \\
\midrule
Claude Opus 4.6        & 0.02          & \bfseries 10.7 & 18.9         & 3.3           \\
Claude Sonnet 4.6      & 0.16          & 17.1          & 31.2          & 4.7           \\
\midrule
GPT-5.4                & 0.29          & 73.1          & 45.0          & 2.0           \\
GPT-5.4 Mini           & 0.21          & 71.8          & 20.0          & 1.0           \\
\midrule
Nova Premier           & 0.33          & 73.7          & 15.0 & 1.1 \\
Nova Pro               & 0.28          & 59.5          & 15.2 & 0.0 \\
\midrule
Qwen 3.5 35B-A3B       & 0.34          & 65.5          & 39.1          & 2.2           \\
Qwen 3.5 2B      & 0.18          & 97.5          & 18.1          & 0.0           \\
Qwen 3.5 4B      & 0.36          & 79.3          & 28.6          & 1.1           \\
Qwen 3.5 9B      & 0.46          & 69.2          & 34.8          & 3.3           \\
Qwen 3.5 27B      & 0.53          & 44.3          & 53.2          & 10           \\

Qwen 3.6 27B      & 0.55          & 77.7          & 32.5          & 5.6           \\
Qwen 3.6 35B-A3B-Q8      & 0.38          & 24.2          & 41.5          & 7.9           \\
\midrule
InternVL 3.5 14B      & 0.01          & 113.3          & 29.2          & 1.1           \\
\multirow{2}{*}{{Pearson correlation coefficients}}  &\multicolumn{4}{c}{$r_{\text{mIoU,MP}} = 0.463$} \\
  & \multicolumn{4}{c}{$r_{\text{RMSE,MP}} = -0.315$}\\ %
\bottomrule
\end{tabular}
\caption{Perception results and key mission metrics. Perception metrics (mIoU, RMSE) are measured on the single-frame proxy task over the 10-mission set. Mission metrics (MP, SR) are measured on the 30-mission test split. All results are averaged over 3 runs. 
}
\label{tab:perception}
\end{table}

To test whether static visual understanding predicts mission success, we evaluate all models on a single-frame perception proxy task that measures object localization (mIoU) and distance estimation (RMSE) on a subset of 10 missions. \Cref{tab:perception} reports the full per-model perception proxy and mission-level results underlying the analysis in \Cref{ssec:proxy}. The extended table confirms that the modest-to-moderate correlation between perception and mission performance holds consistently across all model families, not just the highlighted counterexamples discussed in the main text.

\subsection{Performance Variability Across Mission Runs}
\begin{table}[t!]
\centering
\vspace{2pt}
\footnotesize
\setlength{\tabcolsep}{4pt}
\begin{tabular}{@{}l rr rrr@{}}
\toprule
\textbf{Model}
  & \textbf{SR\,(\%)}$\uparrow$
  & \textbf{MP}$\uparrow$
  & \textbf{OSR\,(\%)}$\uparrow$
  & \textbf{ CR\,(\%)}$\downarrow$
  & \textbf{Eff\,(\%)}$\uparrow$ \\
\midrule
Gemini 3.1 Pro
   & \boldmath $34.8 \pm 13.5$ & \boldmath $73.3 \pm 10.6$
  & $\underline{17.2\pm 7.7}$ & $4.3\pm \text{1.5}$ & $\underline{53.6\pm 10.7}$ \\
Gemini 2.5 Pro
   & $\underline{8.9 \pm 11.5}$  & $\underline{54.2 \pm 20.0}$ & $14.4 \pm 11.5$ & $2.9 \pm \text{1.3}$ & $18.2 \pm \text{15.7}$ \\

Gemini 3 Flash
   & $7.9 \pm \text{9.6}$ & $54.0 \pm \text{19.6}$
  & \boldmath $20.2 \pm 15.8$ & $4.9 \pm \text{4.8}$ & $30.8 \pm \text{14.2}$ \\

Gemini 2.5 Flash
   & $5.6 \pm \text{9.6}$ & $39.0 \pm \text{22.7}$
  & $12.4 \pm \text{11.5}$ & $4.1\pm \text{1.6}$  & $24.7 \pm \text{12.7}$ \\
Gemini 3.1 Flash Lite
  & $6.9 \pm \text{0.0}$ & $24.0 \pm \text{12.3}$ 
  & $8.1 \pm \text{2.0}$ & $2.5 \pm \text{1.2}$ & $29.9 \pm \text{10.3}$ \\

Gemini Robotics 1.5
  & $0.0 \pm \text{0.0}$ & $35.9 \pm \text{14.6}$ 
  & $6.9 \pm \text{4.1}$ & $5.8 \pm \text{4.5}$ & $32.1 \pm \text{16.3}$\\
\midrule
Claude Opus 4.6
   & $3.3 \pm \text{0.0}$ & $18.9 \pm \text{11.5}$
  & $10.0 \pm \text{2.1}$ & $4.7 \pm \text{1.3}$ & $16.6 \pm \text{5.8}$ \\
Claude Sonnet 4.6
   & $4.7 \pm \text{6.6}$ & $31.2 \pm \text{14.3}$
  & $7.1 \pm \text{4.4}$ & $\underline{2.0 \pm 0.6}$ & $30.2 \pm \text{13.6}$ \\
\midrule
GPT-5.4
   & $2.0 \pm \text{0.0}$ & $45.0\pm \text{0.1}$
  & $0.0 \pm \text{0.0}$ & $4.0\pm \text{0.0}$ & $17.9 \pm \text{9.3}$ \\
GPT-5.4 Mini
   & $1.0 \pm \text{2.2}$ & $20.0 \pm \text{8.6}$
  & $0.0 \pm \text{0.0}$ & $5.0 \pm \text{1.7}$ & $25.7 \pm \text{9.5}$ \\
\midrule
Nova Premier
   & $1.1 \pm \text{2.2}$ & $15.0 \pm \text{6.1}$
  & $5.6 \pm \text{4.4}$ & \boldmath $1.7 \pm 1.6$ & $17.0 \pm \text{13.4}$ \\
Nova Pro
   & $0.0 \pm \text{0.0}$ & $15.2 \pm \text{7.6}$
  & $3.3 \pm \text{0.0}$ & $2.0 \pm \text{3.0}$ & \boldmath  $59.0 \pm 16.5$ \\
\midrule
Qwen 3.5 35B-A3B
   & $2.2 \pm \text{0.0}$ & $39.1 \pm \text{12.4}$
  & $7.8 \pm \text{2.2}$ & $7.5\pm \text{2.9}$ & $32.1 \pm \text{14.0}$ \\

\bottomrule
\end{tabular}
\caption{This table reports standard deviations of selected models from \Cref{tab:model_performance}. Results are from 3 identical runs with varying random seeds on the 30-mission test split.}
\label{tab:model_performance_apdx}
\vspace{-4pt}
\end{table}

\label{apdx:std_deviation}
Table~\ref{tab:model_performance_std} reports the standard deviation for each mission metrics (SR, MP, OSR, CR, Eff) measured on the 5-mission Ablation split over 5 experiments with different random seeds. The high standard deviation in SR ($\pm 17.9$) and MP ($\pm 14.9$) for Gemini 3 Flash indicates substantial run-to-run variability. This suggests that single-run evaluation may underestimate or overestimate the model's true performance. This underlines the importance of multi-run evaluation protocols when benchmarking MLLMs on embodied tasks. 
We supplement the results in \Cref{tab:model_performance} with their standard deviations in \Cref{tab:model_performance_apdx}.
\begin{table}[t!]
\centering
\vspace{2pt}
\footnotesize
\setlength{\tabcolsep}{4pt}
\begin{tabular}{@{}l rr rrr@{}}
\toprule
\textbf{Model}
  & \textbf{SR\,(\%)}$\uparrow$
  & \textbf{MP}$\uparrow$
  & \textbf{OSR\,(\%)}$\uparrow$
  & \textbf{ CR\,(\%)}$\downarrow$
  & \textbf{Eff\,(\%)}$\uparrow$ \\
\midrule

Gemini 3 Flash
   & $54.2 \pm \text{17.9}$ & $73.4 \pm \text{14.9}$
  & $45.5 \pm 11.0$ & $0.0 \pm \text{0.0}$ & $29.0 \pm \text{11.0}$ \\

\bottomrule

\end{tabular}
\caption{Standard deviation of mission metrics across 5 runs on Gemini 3 Flash on the 5-Ablation test split.} 
\label{tab:model_performance_std}
\end{table}

\subsection{Quantitative Failure Mode Analysis}
\label{apdx:failure_mode_analysis}

\begin{table}[t]
    \centering
    \begin{tabular}{lrr}
        \hline
        Failure Type & GPT 5.4 & Gemini 3 Flash  \\
        \hline
        Drift and Oscillation & 36.6 & 6.7\\
        Unmet Mission Objectives & 15.9 & 28.9\\
        Altitude / Align Failure & 8.5 & 24.4 \\
        Premature Termination & 7.3 & 14.4 \\
        Continuous Collision & 7.3 & 8.9 \\
        \hline
    \end{tabular}
    \caption{Breakdown of the failure case distribution for two models. The type of mistakes leading to non-successful mission completion and their frequency are highly model-specific. 
    }
    \label{tab:failure_cases}
\end{table}

To better understand why missions fail beyond aggregate metrics such as SR, MP, and Eff, we automatically assign each non-successful episode to a primary failure category using simple rule-based criteria derived from the episode trajectory, terminal state, and reported output. The categories and their assignment criteria are given below, and the resulting failure distributions for two representative models are shown in \Cref{tab:failure_cases}.

\paragraph{Continuous Collision}
Agents with no notion or blatant disregard for operational safety that cause collisions with objects at least three times in a single run fall under this failure mode. 

\paragraph{Drift and Oscillation.} 
Models with low Eff and MP exhaust their step budget without meaningful convergence on the target. 

\paragraph{Premature Termination.}
Models with high Eff, low MP frequently report the mission complete within the first few steps, often before the target is even close. 

\paragraph{Unmet Mission Objectives.} 
This mode corresponds to the OSR $>$ SR gap identified in \Cref{ssec:full_eval}. The agent successfully navigates to the target vicinity but fails to complete the mission objectives (e.g., misidentifying the object, reporting incorrect attributes, not interacting with the target (manipulation missions), or failing to signal completion).

\paragraph{Altitude or Align Failures.} 
The agent neither successfully finished the mission objectives, nor did it navigate somewhere close to the target position ($\| \mathbf{p}_\text{final} - \mathbf{p}^* \| > \tau_d$ or $\tau_\text{final}>\tau_\psi$)

The failure distributions in \Cref{tab:failure_cases} show that error patterns are strongly model-specific rather than universal. GPT~5.4 fails predominantly through \textit{Drift and Oscillation} (36.6\%), indicating that its main weakness is sustained closed-loop control and efficient convergence making that it might not reach the ultimate mission completion stage. In contrast, Gemini~3 Flash shows substantially fewer drift failures (6.7\%) and fails to meet the mission objectives more often (28.9\%). This indicates that it reaches the general target region more often but struggles with precise final positioning, viewpoint selection, or mission completion once there. Gemini~3 Flash also exhibits more \textit{Premature Termination} (14.4\% vs.\ 7.3\%), pointing to a stronger tendency to declare success before sufficient evidence has been gathered. Continuous collision rates are comparatively similar across both models. Overall, these results highlight that embodied MLLM failures arise from different combinations of control, perception, and decision-making errors, which are obscured by aggregate success metrics alone.

None of these patterns is readily visible from static, perception-focused analyses such as those in \Cref{ssec:proxy}. This further supports the need for closed-loop, mission-level evaluation: only such benchmarks can reveal whether current MLLMs fail due to deficient control, poor last-mile perception, premature stopping, or unsafe interaction with the environment.
\subsection{Task-Specific Results}
\label{apdx:Task-Specific Results}
Table~\ref{tab:task_breakdown} breaks down the results by task category, highlighting the relative difficulty of each task for the agents.
\begin{table}[h]
\centering
\vspace{2pt}
\footnotesize
\setlength{\tabcolsep}{8pt}
\begin{adjustbox}{max width=\textwidth}
\begin{tabular}{@{}l ccccc@{}}
\toprule
\textbf{Agent} & \textbf{Overall} & \textbf{Reporting} (17) & \textbf{Inspection} (7) & \textbf{Manipulation} (3) & \textbf{Patrol} (3) \\
\midrule
Gemini 3.1 Pro  & 34.8 / 73.3 & 48.0 / 79.0 & 19.0 / 67.4 &  0.0 / 86.4 & 33.3 / 42.5 \\
Qwen 3.5 27B    & 10.0 / 53.2 & 17.6/56.8 & 0.0/57.1 &  0.0 / 60.6 &  0.0 / 16.4 \\
\midrule
Human Baseline  & 70.0 / 79.0 & 88.9/83.9 & 46.2/77.0 & 60.0 / 96.6 & 16.7 / 43.8 \\
\midrule
Human Baseline (w/ keyboard control) & \humancontrolsr / \humancontrolmp & 92.13/98.5 & 71.4/95.0 & 66.7/95.5 & 88.9/67.23 \\
\bottomrule
\end{tabular}
\end{adjustbox}
\caption{Breakdown of task-specific performance. The table shows the results for visual inspection, manipulation, and patrol tasks separately, with the number of missions for each in parentheses, presented as SR / MP format.}
\label{tab:task_breakdown}
\vspace{-4pt}
\end{table}

\section{Additional Benchmark Details}
\label{apdx:details_on_benchmark_generation}

\subsection{Simulation Environment.}
We adopt CosysAirSim~\citep{jansen2023cosys} as our simulation setup for two reasons. First, CosysAirSim is a community-maintained fork of AirSim~\citep{shah2017airsim}, which has been actively adopted in the drone research literature. Second, CosysAirSim supports Unreal Engine 5~\citep{unrealengine}, allowing us to easily modify the environment assets within the UE5 editor. An alternative is UnrealCV~\citep{qiu2017unrealcv}, which also provides programmatic access to Unreal Engine.

Using this setup, we developed or adopted five distinct environments for \benchmark{}. The \emph{Forest} environment was built from scratch using free assets from the Unreal Fab store. The \emph{Neighborhood (NH)} environment was developed by extending a small purchased scene, and the \emph{City} environment was similarly built on top of an existing base. The \emph{AirSim Neighborhood (AirSimNH)} and \emph{Savannah} environments were sourced from pre-compiled binaries distributed with the official AirSim release.

\subsection{Mission Definition.}

Initial mission concepts and ideas inspired by real-world operational scenarios were first conceived and collected. Based on these plans, relevant assets were placed at appropriate locations within each environment, and a compiled Linux binary of each scene was packaged. A considerable amount of effort went into this stage: unlike object navigation tasks, where reaching the vicinity of a target object suffices and data collection is comparatively straightforward, mission planning requires environments to be deliberately designed so that the intended real-world tasks can be empirically evaluated. 
Therefore, all mission instructions were written by humans rather than generated by LLMs, with the goal of reflecting realistic UAV operator commands, including the mild ambiguity and underspecification often present in practice. Each instruction and mission setup was reviewed by multiple authors for validity, consistency, and solvability.

For each mission, a fixed start pose and position was set such that the target or target region is at least partially visible while still requiring deliberate planning and execution for solving the mission. 
For reporting, inspection, and manipulation missions, the initial view may already contain the relevant object or area, but successful completion still requires further approach and alignment. For patrol missions, the initial view reveals only part of the route or region, preventing trivial completion from a single observation.
This initial state is reused for both reference-data collection as well as model evaluation.

\subsection{Reference Data Collection}

Once the packaged environments were ready, data collection commenced. Reference trajectories were obtained by manually piloting a UAV in the simulator while following the mission instruction. The purpose of this process was not to imitate the model input interface, but to obtain an executable reference trajectory and a valid terminal state for each mission. AirSim's built-in recording functionality captured time-stamped poses and images, producing a log file alongside the corresponding frames. %

Using this pipeline, a total of 120 missions were collected. Each collected mission was then manually annotated with additional mission-specific ground truth information in form of a reference caption specifying the expected textual mission report $\mathcal{R}$ for reporting missions.

\subsection{Quality Assurance and Iterative Refinement.}
To ensure a high quality of \benchmark{} and realistic and solvable missions, multiple cycles of environment design, packaging, and mission generation were conducted. During post-packaging evaluation and manual test flights, visual artifacts, simulation inconsistencies, or opportunities for improvement were occasionally identified and resolved.%

\subsection{Evaluation Metrics for Patrol}
\label{apdx:evaluation_metric}
\begin{figure}[t]
    \centering
    \includegraphics[width=0.6\textwidth]{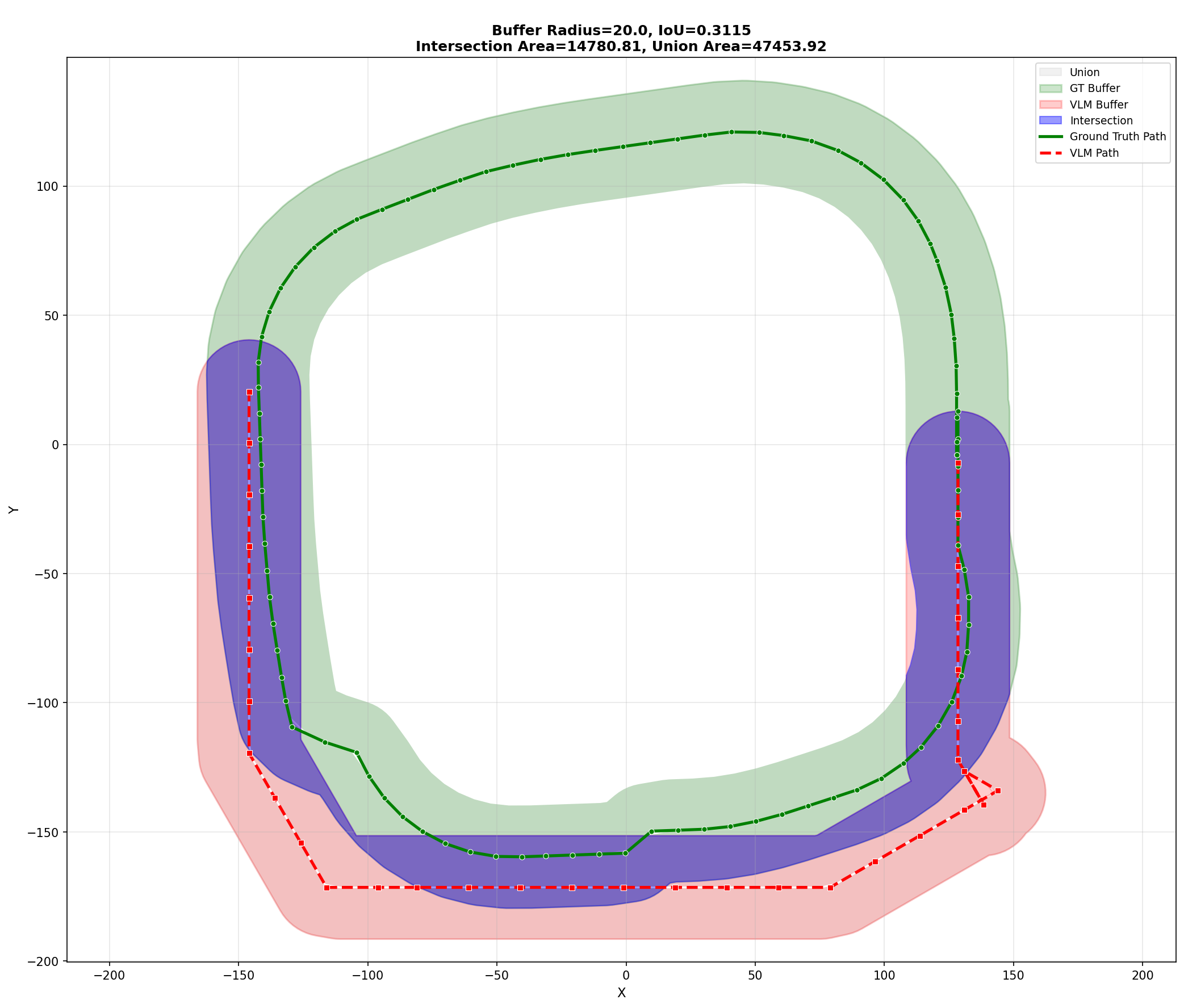}
    \caption{\textbf{Patrol IoU evaluation.} The ground-truth route (green) and predicted route (red) are each dilated by a $20\,\text{m}$ buffer. The shaded overlap region (purple) determines $\text{IoU}_{\text{patrol}}$; here the agent exhausts its step budget before completing the intended route, resulting in a low mIoU.}
    \label{fig:patrol_iou}
\end{figure}
Patrol missions require the agent to follow a closed or open route that covers a prescribed area. Since comparing two trajectories point-wise is sensitive to speed and temporal alignment, we instead adopt a spatial overlap measure. We first uniformly resample both the ground-truth trajectory $\tau^*$ and the predicted trajectory $\hat{\tau}$ into $N{=}100$ segments, then dilate each with a buffer of radius $r{=}20\,\text{m}$, producing two planar regions $\mathcal{B}(\tau^*, r)$ and $\mathcal{B}(\hat{\tau}, r)$. The metric is the Intersection-over-Union of these regions:

\begin{equation}
  \text{IoU}_{\text{patrol}} = \frac{|\,\mathcal{B}(\tau^*,\, r) \;\cap\; \mathcal{B}(\hat{\tau},\, r)\,|}{|\,\mathcal{B}(\tau^*,\, r) \;\cup\; \mathcal{B}(\hat{\tau},\, r)\,|}
\end{equation}

A patrol is considered successful when $\text{IoU}_{\text{patrol}} > 0.5$. The buffer radius mitigates minor lateral deviations while still penalizing trajectories that skip or shortcut large portions of the intended route, as depicted in \Cref{fig:patrol_iou}. The mean IoU (mIoU) across patrol scenarios can also be utilized as continuous performance indicator.

\subsection{LLM-as-judge validation}
\label{apdx:llm_as_judge_details}
We utilize Gemini-3-flash-preview to compare the expected GT and predicted model report and to conclude whether they ``softly'' match (i.e., despite minor wording/spacing/formatting differences) as the goal is to test whether the UAV reached a suitable viewpoint and extracted the mission-relevant information, not whether the model matches a brittle string format.
To ensure the reliability of this LLM-as-judge setup, we manually verify 150 randomly selected runs. Exact string matching rejects many correct reports due to formatting or semantic equivalence (e.g., solar-powered vs. solar powered,  recyclable waste vs. recyclables). The LLM-as-judge has near-perfect agreement with independent human correctness assessments and only differs by more strictly assessing close gt-prediction mismatches in three cases (e.g., rejecting ``red and white antennas'' for GT ``cell tower'' while human accepts it). When replacing the LLM-as-judge with exact matching, the success rate drops from 34.8 to 19.1 for Gemini 3.1 Pro, significantly under-counting successful mission completion. Table~\ref{tab:llm_as_judge} compares the alignment with human assessments of the LLM-as-judge against a simple exact match. Section~\ref{apdx:llm_judge_prompt} reports the prompt used for the LLM judge.
\begin{table}[h]
\centering
\vspace{2pt}
\footnotesize
\setlength{\tabcolsep}{12pt}
\begin{tabular}{@{}l rr@{}}
\toprule
\textbf{Evaluator} & \textbf{Precision} & \textbf{Recall} \\
\midrule
LLM-as-judge  & 1.00 & 0.94 \\
Exact match   & 1.00 & 0.44 \\
\bottomrule
\end{tabular}
\caption{Validation of LLM-as-judge against exact matching across 150 human-annotated report evaluation cases. Exact match artificially suppresses success rates due to minor formatting discrepancies, whereas the LLM-as-judge closely mirrors human assessment of semantic correctness.}
\label{tab:llm_as_judge}
\vspace{-4pt}
\end{table}

\section{Model Details}
\label{apdx:model_details}

\subsection{Inference Parameters}
All models were run with their default configurations, and for certain open models, we additionally applied the settings recommended by their respective inference providers.
For Qwen3.6-35B-A3B-Q8 is the 8-bit quantized version of the model, we used instruct-mode parameters: temperature=1.0, top\_p=0.95, top\_k=20, min\_p=0.0, presence\_penalty=1.5, and repetition\_penalty=1.0.

For the Gemini models, the default dynamic thinking levels were applied to Gemini 2.5 Flash, Gemini 2.5 Pro. Gemini Robotics 1.6 used a "high" thinking level. For the Gemini 3 models, the default ``high'' thinking level, which corresponds to dynamic thinking, was used.
\subsection{Inference Cost}
\begin{table}
\centering
\begin{tabular}{lccc}
\hline
\textbf{Model Name} & \textbf{Input (\$/M tokens)} & \textbf{Output (\$/M tokens)} & \textbf{Total Cost (\$)} \\
\hline
Gemini 3.1 Pro & 4 & 18 & 56.9 \\

Claude Opus 4.6 & 5 & 25 & 71.9 \\

GPT-5.4  & 2.5 & 15 & 36.7 \\

Nova Pro  & 1.25 & 10 & 19.2 \\

\hline
\end{tabular}
\caption{Per-model API cost for the full Test split evaluation.}
\label{apdx:cost_table}
\end{table}

\Cref{apdx:cost_table} summarizes the per-model API cost for evaluating the full Test split (30 missions $\times$ 3 runs). Estimates assume approximately 12.8M input tokens and 0.3M output tokens per model, based on three egocentric images per step at 1920$\times$1080 resolution. Open-weight models (e.g., Qwen) are excluded as they were run on local compute.

\begin{wrapfigure}{r}{0.25\textwidth}
\centering
\vspace{-12mm}
\includegraphics[width=\linewidth]{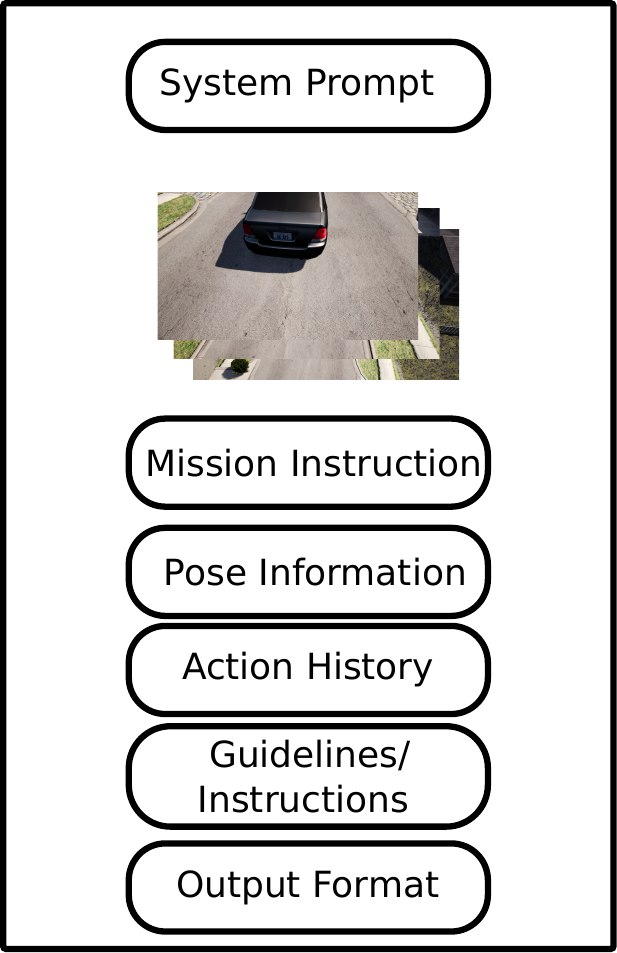}
\caption{
Structure of model input (prompts)
}
\label{fig:prompt_structure}
\end{wrapfigure}
\subsection{Model Prompts}
\label{apdx:prompt_details}
We provide the raw prompts used in \benchmark{} in the following. The structure of the model input is also visualized in \Cref{fig:prompt_structure}. The system prompt specifies the agent role, available action primitives, and required output fields (bounding box, done flag, reasoning). The per-step prompt includes the current observation, a two-frame history buffer, and the mission instruction. %
Placeholders that will be replaced with mission-specific or observational information at inference time are indicated in \colorbox{blue!10}{blue}. 
All prompts are handcrafted and consider common failure cases across model families. As previously mentioned in \Cref{apdx:limitations}, no extensive, model-specific prompt engineering was conducted with the design objective being good overall zero-shot generalization for any MLLM. 

\paragraph{System Prompt.}
\VerbatimInput[
  breaklines=true,
  breakanywhere=true,
  fontsize=\tiny,
  numbers=left,
  frame=single,
  commandchars=+\[\]
]{contents/sections/prompts/system_prompt.txt}

\paragraph{Per-Step Prompt.}
\VerbatimInput[
  breaklines=true,
  breakanywhere=true,
  fontsize=\tiny,
  numbers=left,
  frame=single,
  commandchars=+\[\]
]{contents/sections/prompts/user_prompt.txt}

\paragraph{Perception Proxy Task Prompt.}
\VerbatimInput[
  breaklines=true,
  breakanywhere=true,
  fontsize=\tiny,
  numbers=left,
  frame=single,
  commandchars=+\[\]
]{contents/sections/prompts/perception_prompt.txt}

\paragraph{LLM as a judge prompt}
\label{apdx:llm_judge_prompt}
\VerbatimInput[
  breaklines=true,
  breakanywhere=true,
  fontsize=\tiny,
  numbers=left,
  frame=single,
  commandchars=+\[\]
]
{contents/sections/prompts/evaluation_prompt.txt}

\end{document}